\documentclass{article}

\usepackage[final,nonatbib]{neurips_2024}

\usepackage[utf8]{inputenc} 
\usepackage[T1]{fontenc}    
\usepackage{url}            
\usepackage{booktabs}       
\usepackage{amsfonts}       
\usepackage{nicefrac}       
\usepackage{microtype}      
\usepackage{xcolor}         
\usepackage{graphicx}

\definecolor{myblue}{rgb}{0.21,0.49,0.74}
\usepackage[pagebackref,breaklinks,colorlinks,citecolor=myblue]{hyperref}

\usepackage{xspace}
\usepackage{pifont}
\usepackage{xfrac}
\usepackage{multirow}
\usepackage{makecell}
\usepackage{colortbl}
\usepackage{titletoc}
\usepackage{amsmath}
\usepackage{wrapfig}
\usepackage[capitalise]{cleveref}
\crefname{section}{Sec.}{Secs.}
\Crefname{section}{Section}{Sections}
\Crefname{table}{Table}{Tables}
\crefname{table}{Tab.}{Tabs.}

\makeatletter
\DeclareRobustCommand\onedot{\futurelet\@let@token\@onedot}
\def\@onedot{\ifx\@let@token.\else.\null\fi\xspace}

\def\eg{\emph{e.g}\onedot}

\def\ie{\emph{i.e}\onedot}

\makeatother

\newcommand{\modelname}{Vista\xspace}

\title{\textit{\modelname}: A Generalizable Driving World Model with \\ High Fidelity and Versatile Controllability}

\author{
  Shenyuan Gao$^{1,2}$\quad Jiazhi Yang$^{2}$\quad Li Chen$^{2,5}$\quad Kashyap Chitta$^{3,4}$\quad Yihang Qiu$^{2}$ \\
  \textbf{Andreas Geiger}$^{3,4\dagger}$\quad \textbf{Jun Zhang}$^{1\dagger}$\quad \textbf{Hongyang Li}$^{2,5\dagger}$ \\[2mm]
  $^{1}$~Hong Kong University of Science and Technology \quad 
  $^{2}$~OpenDriveLab at Shanghai AI Lab  \\
   $^3$~University of Tübingen\quad $^4$~Tübingen AI Center \quad $^{5}$~University of Hong Kong\\[2mm]
  Code and model: \href{https://github.com/OpenDriveLab/Vista}{\texttt{github.com/OpenDriveLab/Vista}} \\
  Demo page: \href{https://opendrivelab.com/Vista/}{\texttt{opendrivelab.com/Vista}}
  \vspace{-10mm}
}

\begin{document}

\maketitle

\enlargethispage{2\baselineskip}
{\let\thefootnote \relax \footnote{Primary contact to Shenyuan at \texttt{sygao@connect.ust.hk} ~~$^\dagger$Equal advising.}}

\setcounter{footnote}{0}

\begin{abstract}
World models can foresee the outcomes of different actions, which is of paramount importance for autonomous driving. Nevertheless, existing driving world models still have limitations in generalization to unseen environments, prediction fidelity of critical details, and action controllability for flexible application. In this paper, we present \textit{\modelname}, a generalizable driving world model with high fidelity and versatile controllability. Based on a systematic diagnosis of existing methods, we introduce several key ingredients to address these limitations. To accurately predict real-world dynamics at high resolution, we propose two novel losses to promote the learning of moving instances and structural information. We also devise an effective latent replacement approach to inject historical frames as priors for coherent long-horizon rollouts. For action controllability, we incorporate a versatile set of controls from high-level intentions (command, goal point) to low-level maneuvers (trajectory, angle, and speed) through an efficient learning strategy. After large-scale training, the capabilities of \modelname can seamlessly generalize to different scenarios. Extensive experiments on multiple datasets show that \modelname outperforms the most advanced general-purpose video generator in over $70\%$ of comparisons and surpasses the best-performing driving world model by $55\%$ in FID and $27\%$ in FVD. Moreover, for the first time, we utilize the capacity of \modelname itself to establish a generalizable reward for real-world action evaluation without accessing the ground truth actions.
\end{abstract}

\section{Introduction}
\label{sec:intro}
Driven by scalable learning techniques, autonomous driving has made encouraging strides over the past few years~\cite{chen2023end,hu2023planning,yan2024forging}. However, intricate and out-of-distribution situations are still intractable for state-of-the-art techniques~\cite{li2023ego}. One promising solution lies in world models~\cite{hu2023toward,lecun2022path}, which infer the possible future states of the world from historical observations and alternative actions, in turn assessing the feasibility of such actions. They hold the potential to reason with uncertainty and avoid catastrophic errors~\cite{hu2023gaia,lecun2022path,wang2023driving}, thereby promoting generalization and safety in autonomous driving.

\begin{figure}[t!]
\centering
\includegraphics[width=0.95\textwidth]{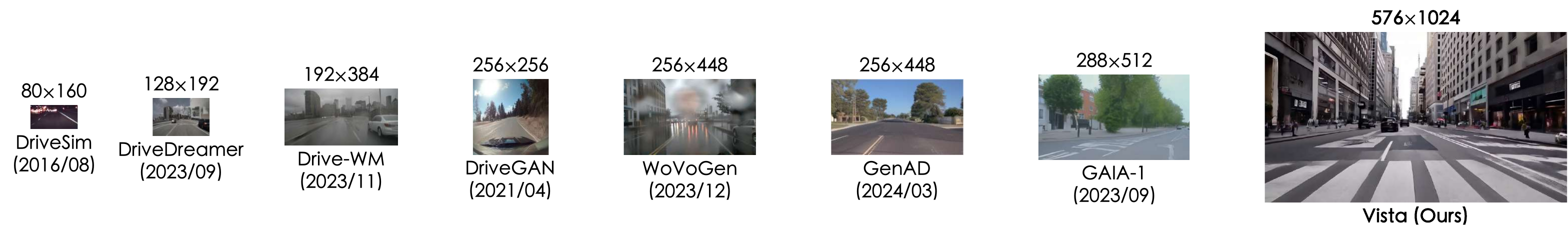}
\vspace{-3mm}
\caption{\textbf{Resolution comparison.} \modelname predicts at a higher resolution than previous literature.}
\label{fig:resolution}
\vspace{-3mm}
\end{figure}

Although a primary prospect of world models is to enable the generalization ability to novel environments, existing driving world models are still constrained by data scale~\cite{lu2023wovogen,wang2023drivedreamer,wang2023driving,zhang2023learning,zheng2023occworld} and geographical coverage~\cite{hu2023gaia,jia2023adriver}. As summarized in \Cref{tab:vwm} and \cref{fig:resolution}, they are also often confined to low frame rates and resolutions, resulting in a loss of critical details. Furthermore, most models only support a single control modality such as the steering angle and speed. This is insufficient to express various action formats ranging from high-level intentions to low-level maneuvers, and incompatible with the outputs of prevalent planning algorithms~\cite{casas2021mp3,chen2020learning,chitta2022transfuser,hu2022st,hu2023planning,jiang2023vad}. In addition, generalizing action controllability to unseen datasets is understudied. These limitations impede the applicability of existing works, making it imperative to develop a world model that overcomes these limitations.

To this end, we introduce \textit{\modelname}, a driving world model that is proficient in cross-domain generalization, high-fidelity prediction, and multi-modal action controllability. Specifically, we develop the predictive model on a large corpus of worldwide driving videos~\cite{yang2024generalized} to foster its generalization ability. To enable coherent future extrapolation, we condition \modelname on three essential dynamic priors (\cref{sec:dyamics}). Instead of solely relying on the standard diffusion loss~\cite{blattmann2023stable}, we introduce two explicit loss functions to enhance dynamics and preserve structural details (\cref{sec:dyamics}), promoting \modelname's ability to simulate realistic futures at high resolution. For flexible controllability, we incorporate a versatile set of action formats, including both high-level intentions such as commands and goal points, as well as low-level maneuvers like trajectories, steering angles, and speeds. These action conditions are injected via a unified interface, which is learned through an efficient training strategy (\cref{sec:action}). Consequently, as \cref{fig:teaser} shows, \modelname acquires the ability to anticipate realistic futures at 10 Hz and 576$\times$1024 pixels, and obtains versatile action controllability across various levels of granularity. We also demonstrate the potential of \modelname as a generalizable reward function to evaluate the reliability of different actions.

\begin{table}[t!]
\caption{\textbf{Real-world driving world models.} Trained on large-scale high-quality driving data, \modelname performs at high spatiotemporal resolution and supports versatile action controllability. \colorbox[gray]{0.9}{Private data}.}
\label{tab:vwm}
\vspace{-2mm}
\centering
\scalebox{0.7}{
\begin{tabular}{p{0.20\textwidth} | >{\centering}p{0.14\textwidth} >{\centering}p{0.14\textwidth } >{\centering}p{0.14\textwidth} | >{\centering}p{0.12\textwidth} >{\centering}p{0.12\textwidth} >{\centering}p{0.12\textwidth} >{\centering\arraybackslash}p{0.12\textwidth}}
\toprule
\multicolumn{1}{l|}{\multirow{2}{*}{\textbf{Method}}} & \multicolumn{3}{c|}{\textbf{Model Setups}} & \multicolumn{4}{c}{\textbf{Action Control Modes}} \\
& Data Scale & Frame Rate & Resolution & Angle\&Speed & Trajectory & Command & Goal Point \\
\midrule
DriveSim~\cite{santana2016learning} & 7h & 5 Hz & 80$\times$160 & \ding{51} & & & \\
DriveGAN~\cite{kim2021drivegan} & \cellcolor[gray]{0.9}160h & 8 Hz & 256$\times$256 & \ding{51} & & & \\
DriveDreamer~\cite{wang2023drivedreamer} & 5h & 12 Hz & 128$\times$192 & \ding{51} & & & \\
Drive-WM~\cite{wang2023driving} & 5h & 2 Hz & 192$\times$384 & & \ding{51} & & \\
WoVoGen~\cite{lu2023wovogen} & 5h & 2 Hz & 256$\times$448 & \ding{51} & & & \\
ADriver-I~\cite{jia2023adriver} & \cellcolor[gray]{0.9}300h & 2 Hz & 256$\times$512 & & & \ding{51} & \\
GenAD~\cite{yang2024generalized} & 2000h & 2 Hz & 256$\times$448 & & \ding{51} & \ding{51} & \\
GAIA-1~\cite{hu2023gaia} & \cellcolor[gray]{0.9}4700h & 25 Hz & 288$\times$512 & \ding{51} & & & \\
\midrule
\modelname (Ours) & 1740h & 10 Hz & 576$\times$1024 & \ding{51} & \ding{51} & \ding{51} & \ding{51} \\
\bottomrule
\end{tabular}
}
\vspace{-3mm}
\end{table}

Our contributions are three-fold: \textbf{(1)}~We present \textit{\modelname}, a generalizable driving world model that can predict realistic futures at high spatiotemporal resolution. Its prediction fidelity is greatly improved by two novel losses that capture dynamics and preserve structures, along with exhaustive dynamic priors to sustain consistency in long-horizon rollouts. \textbf{(2)}~Propelled by an efficient learning strategy, we integrate versatile action controllability into \modelname through a unified conditioning interface. The action controllability of \modelname can also generalize to different domains in a zero-shot manner. \textbf{(3)}~We conduct comprehensive experiments across multiple datasets to verify the effectiveness of \modelname. It outperforms the most competitive general-purpose video generator and sets a new state-of-the-art on nuScenes. Our empirical evidence shows that \modelname can be used as a reward function to assess actions.

\section{Preliminary}
\label{sec:preliminary}
We initialize \modelname with the pretrained Stable Video Diffusion (SVD)~\cite{blattmann2023stable}, a latent diffusion model for image-to-video generation. For sampling flexibility, SVD adopts a continuous-timestep formula~\cite{karras2022elucidating,song2020score}. It converts data samples $\boldsymbol{x}$ to noise $\boldsymbol{n}$ through a diffusion process $p(\boldsymbol{n}|\boldsymbol{x})\sim\mathcal N(\boldsymbol{x},\sigma^{2}\mathbf{I})$, and generates new samples by progressively denoising the latent towards $\sigma_{0}=0$ from Gaussian noise. The training of SVD can be simplified to minimizing $\mathbb{E}_{\boldsymbol{x},\sigma,\boldsymbol{n}}\Big[\lambda_{\sigma}\Vert D_{\theta}(\boldsymbol{n};\sigma)-\boldsymbol{x}\Vert^{2}\Big]$, where $D_{\theta}$ is a parameterized UNet denoiser and $\lambda_{\sigma}$ is a re-weighting function omitted hereinafter for brevity. Based on this framework, SVD processes a sequence of noisy latent $\boldsymbol{n}=\{n_{1},n_{2},...,n_{K}\}\in\mathbb{R}^{K\times C\times H\times W}$ and generates a video with $K=25$ frames. The generation process is guided by a condition image, whose latent is concatenated channel-wise to the inputs, serving as a reference for content generation.

\begin{figure}[t!]
\centering
\includegraphics[width=0.99\textwidth]{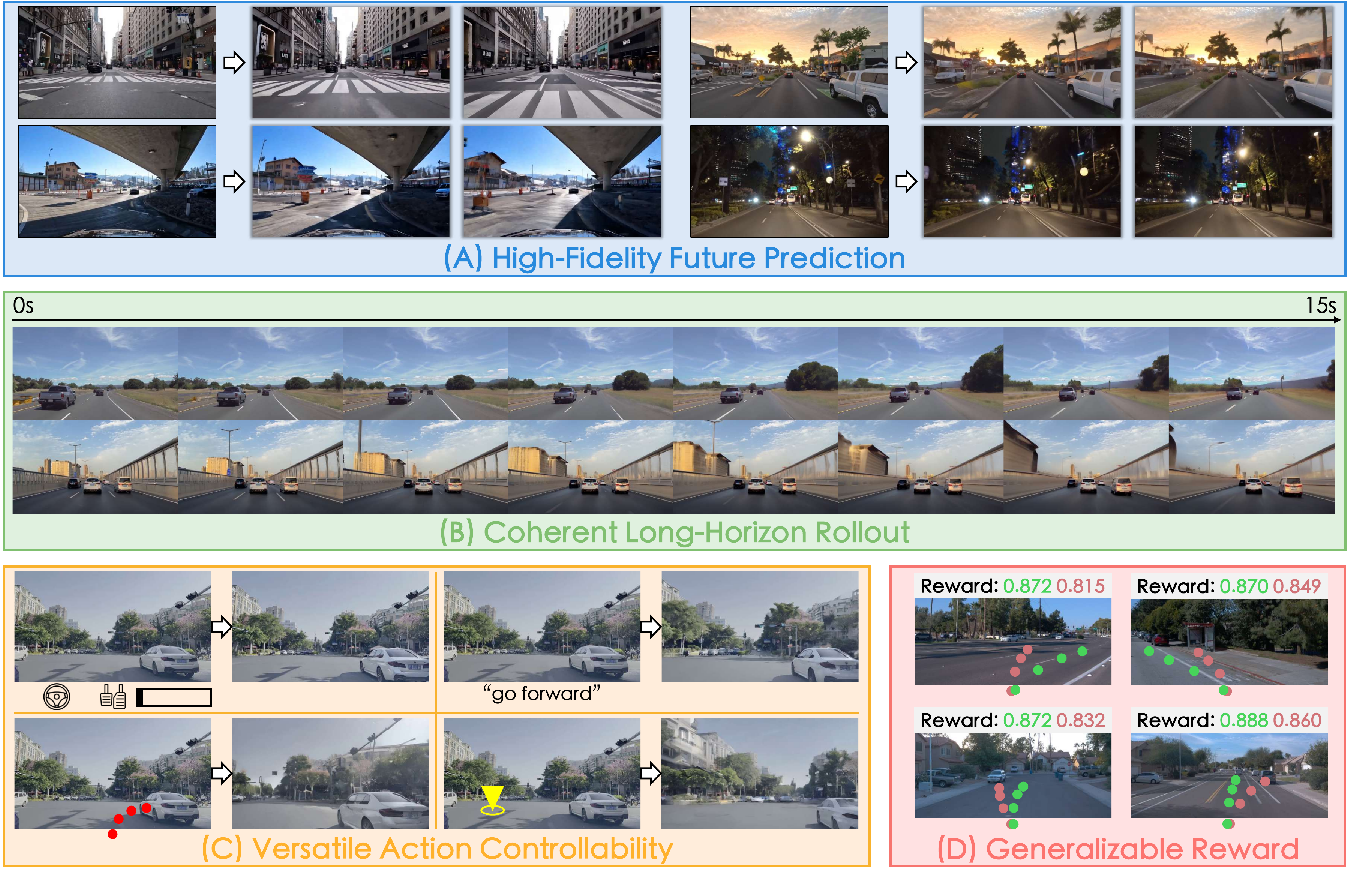}
\vspace{-3mm}
\caption{\textbf{Capabilities of \modelname.} Starting from arbitrary environments, \modelname can anticipate realistic and continuous futures at high spatiotemporal resolution~\textbf{(A-B)}. It can be controlled by multi-modal actions~\textbf{(C)}, and serve as a generalizable reward function to evaluate real-world driving actions~\textbf{(D)}.}
\label{fig:teaser}
\vspace{-3mm}
\end{figure}

Despite the high aesthetic quality, SVD lacks several key properties to function as a driving world model. As shown in \cref{sec:experiments}, the first frame predicted by SVD is not identical to the condition image, making it impractical for autoregressive rollout due to content inconsistency. In addition, SVD struggles with the intricate dynamics of driving scenarios, entailing implausible motions. Moreover, SVD cannot be controlled by any action format. In contrast, we aim to build a generalizable driving world model that predicts high-fidelity futures with realistic dynamics. It ought to be continuously extendable to long horizons and flexibly controllable by multi-modal actions as illustrated in \cref{fig:teaser}.

\section{Learning a Generalizable Driving World Model}
\label{sec:method}
As depicted in \cref{fig:pipeline}, \modelname adopts a two-phase training pipeline. First, we build a dedicated predictive model, which involves a latent replacement approach to enable coherent future prediction and two novel losses to enhance fidelity (\cref{sec:dyamics}). To ensure the generalization to unseen scenarios, we utilize the largest public driving dataset~\cite{yang2024generalized} for training. In the second phase, we incorporate multi-modal actions to learn action controllability with an efficient and collaborative training strategy (\cref{sec:action}). Using the ability of \modelname, we further introduce a generalizable approach to evaluate actions (\cref{sec:reward}).

\subsection{Phase One: Learning High-Fidelity Future Prediction}
\label{sec:dyamics}
\noindent\textbf{Basic Setup.}
Since world models are initiated to predict futures from the current state, the starting of their prediction should be firmly aligned with the condition image. Therefore, we tailor SVD into a dedicated predictive model by imposing the first frame as the condition image and discarding the noise augmentation~\cite{blattmann2023stable,ho2022cascaded} during training. With this prediction ability, \modelname can perform long-term rollouts by iteratively predicting short-term clips and resetting the condition image with the last clip.

\noindent\textbf{Dynamic Prior Injection.}
Nevertheless, using the aforementioned setup for training often results in irrational dynamics with respect to historical frames, especially in long-term rollouts. We conjecture that this mainly arises from the ambiguity caused by insufficient priors about the tendency of future motions, which is also a common limitation of existing driving world models~\cite{hu2023gaia,kim2021drivegan,wang2023drivedreamer,wang2023driving,yang2024generalized}.

\begin{figure}[t!]
\centering
\includegraphics[width=0.99\textwidth]{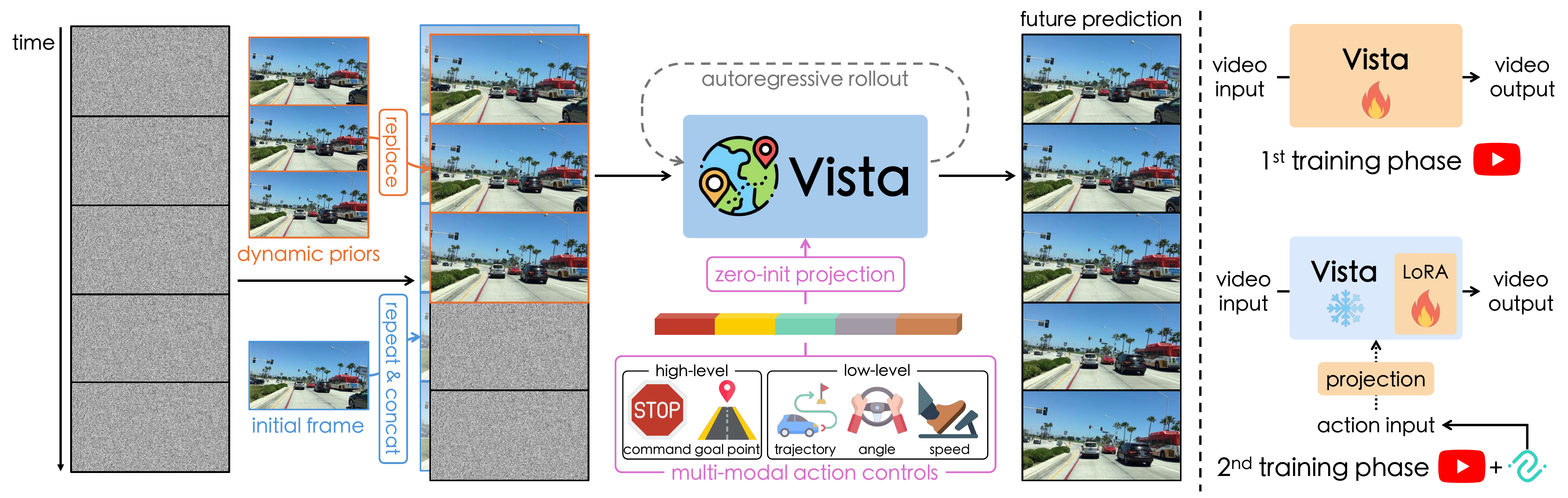}
\vspace{-2mm}
\caption{\textbf{[Left]: \modelname pipeline.} In addition to the initial frame, \modelname can absorb more priors about future dynamics via latent replacement. Its prediction can be controlled by different actions and be extended to long horizons through autoregressive rollouts. \textbf{[Right]: Training procedure.} \modelname takes two training phases, where the second phase freezing the pretrained weights to learn action controls.}
\label{fig:pipeline}
\vspace{-2mm}
\end{figure}

Estimating coherent futures requires at least three essential priors that inherently govern the future motion of instances in the scene: position, velocity, and acceleration. Since velocity and acceleration are the first- and second-order derivative of position respectively, these priors can be entirely derived by using three consecutive frames for conditioning. Concretely, we build a frame-wise mask $\boldsymbol{m}\in\{0,1\}^{K}$ with a length of $K$ to indicate the presence of condition frames. The mask is set sequentially following the time order, with at most three elements being assigned as $1$ to denote three condition frames. Instead of concatenating additional channels to the inputs, we inject new condition frames by replacing the corresponding noisy latent $n_{i}$ with the clean latent $z_{i}$ encoded by the image encoder. Formally, the input latent is constructed as $\hat{\boldsymbol{n}}=\boldsymbol{m}\cdot\boldsymbol{z}+(1-\boldsymbol{m})\cdot\boldsymbol{n}$ (see \cref{fig:pipeline} [Left]). To discern the clean latent, we duplicate a new timestep embedding from the pretrained weights and allocate it to the condition frames according to $\boldsymbol{m}$. The timestep embeddings for condition frames and prediction frames are trained separately. Compared to channel-wise concatenation, we find that replacing the latent is more effective and flexible in absorbing varying numbers of condition frames. In addition, we observe that the replaced latent, when applied to SVD directly, does not degrade its generation quality. Thus, the original performance will not be disturbed when the training is launched. Since there is no need to predict the observed condition frames, we exclude them from the loss as follows:
\begin{equation}
\label{eq:diffusion}
\mathcal{L}_{\text{diffusion}}=\mathbb{E}_{\boldsymbol{z},\sigma,\hat{\boldsymbol{n}}}\Big[\sum^{K}_{i=1}(1-m_{i})\odot\Vert D_{\theta}(\hat{n}_{i};\sigma)-z_{i}\Vert^{2}\Big],
\end{equation}
where $D_{\theta}$ is the UNet denoiser that shares the same architecture with SVD. With the replaced latent holding sufficient priors, \modelname can fully capture the status of the surrounding instances and predict more coherent and plausible long-term futures through iterative rollouts. In practice, we leverage the last three frames of a predicted clip as dynamic priors for the next prediction step during rollouts.

\begin{figure}[t!]
\centering
\includegraphics[width=0.95\textwidth]{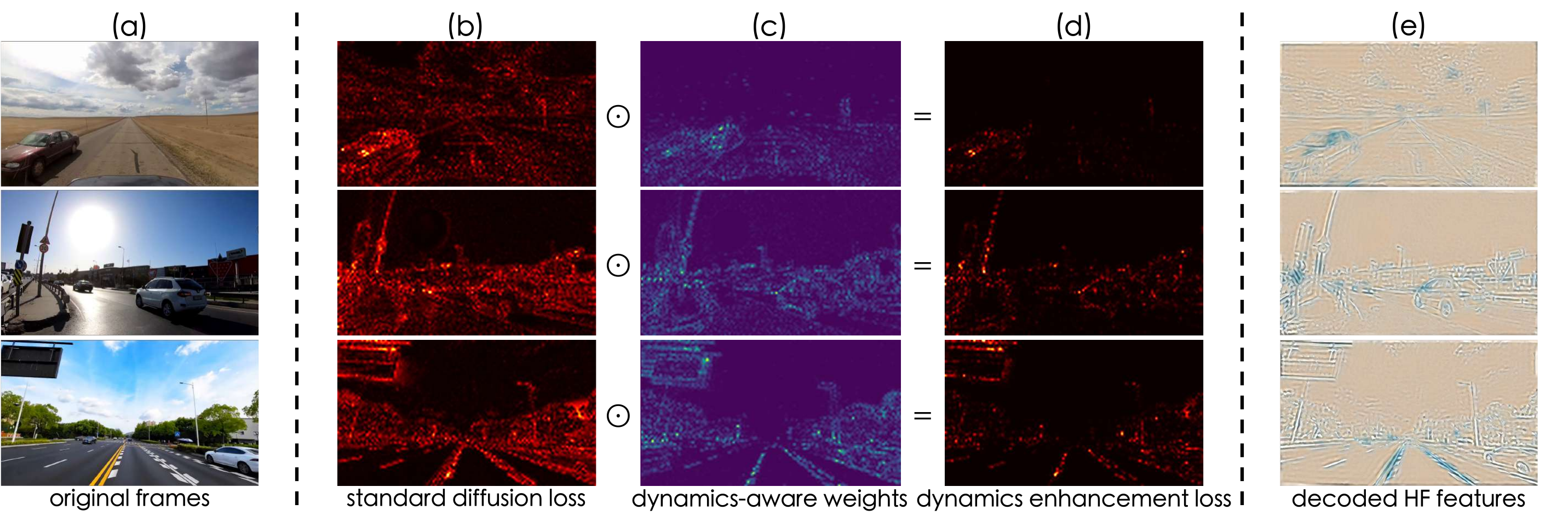}
\vspace{-3mm}
\caption{\textbf{Illustration on loss design.} Different from the standard diffusion loss~\textbf{(b)} that is distributed uniformly, our dynamics enhancement loss~\textbf{(d)} enables an adaptive concentration on critical regions~\textbf{(c)} (\eg, moving vehicles and roadsides) for dynamics modeling. Moreover, by explicitly supervising high-frequency features~\textbf{(e)}, the learning of structural details (\eg, edges and lanes) can be enhanced.}
\label{fig:loss_vis}
\vspace{-3mm}
\end{figure}

\noindent\textbf{Dynamics Enhancement Loss.}
Unlike general videos that cover rather small spaces, driving videos capture much larger scenes~\cite{yang2024generalized}. In most driving videos, distant and monotonous regions dominate the view, with the moving foreground instances only occupying a relatively small area~\cite{chen2023geodiffusion}. However, the latter often exhibit higher stochasticity, complicating their prediction. Since \cref{eq:diffusion} supervises all outputs uniformly, it cannot effectively discriminate the nuances of different regions as \cref{fig:loss_vis}(b) shows. As a result, the model cannot efficiently learn to predict realistic dynamics in crucial regions.

As the discrepancy between two adjacent frames provides considerable motion patterns~\cite{wang2023recipe,wu2023policy}, we introduce an additional supervision to encourage the learning of dynamics for crucial regions. To be specific, we first introduce a dynamics-aware weight $\boldsymbol{w}=\{w_{2},w_{3},...,w_{k}\}\in\mathbb{R}^{K-1\times C\times H\times W}$ that highlights the regions where the prediction has inconsistent motion compared to the ground truth:
\begin{equation}
\label{eq:weight}
w_{i}=\Vert (D_{\theta}(\hat{n}_{i};\sigma)-D_{\theta}(\hat{n}_{i-1};\sigma))-(z_{i}-z_{i-1})\Vert^{2}.
\end{equation}
For numerical stability, we normalize $\boldsymbol{w}$ within each video clip. As shown in \cref{fig:loss_vis}(c), the weight amplifies the presence of large motion disparities, highlighting dynamic regions while excluding monotonous backgrounds. Given the causality of future prediction, \ie subsequent frames ought to follow previous ones, we define a new loss by penalizing the latter frame of each adjacent frame pair:
\begin{equation}
\label{eq:dynamics}
\mathcal{L}_{\text{dynamics}}=\mathbb{E}_{\boldsymbol{z},\sigma,\hat{\boldsymbol{n}}}\Big[\sum^{K}_{i=2}\texttt{sg}(w_{i})\odot(1-m_{i})\odot\Vert D_{\theta}(\hat{n}_{i};\sigma)-z_{i}\Vert^{2}\Big],
\end{equation}
where $\texttt{sg}(\cdot)$ stops the gradient. By adaptively re-weighting the standard diffusion loss, $\mathcal{L}_{\text{dynamics}}$ can boost the learning efficiency of dynamic regions, \eg, the moving vehicles and sidewalks in \cref{fig:loss_vis}(d).

\noindent\textbf{Structure Preservation Loss.}
The trade-off between perceptual quality and motion intensity has been widely acknowledged in video generation~\cite{bar2024lumiere,girdhar2023emu,kondratyuk2023videopoet,zhang2023i2vgen}, and our case is no exception. When it comes to high-resolution prediction for dynamic driving scenarios, we discover that the predicted structural details degrade severely with over-smoothed or broken objects, \eg, the outlines of vehicles unravel quickly as they move (see \cref{fig:loss_effect}). To alleviate this problem, it is important to place more emphasis on structural details. Based on the fact that structural details, such as edges and textures, mainly reside in high-frequency components, we identify them in the frequency domain as follows:
\begin{equation}
\label{eq:hpf}
z'_{i}=\mathcal{F}(z_{i})=\texttt{IFFT}\big(\mathcal{H}\odot\texttt{FFT}(z_{i})\big),
\end{equation}
where $\texttt{FFT}$ and $\texttt{IFFT}$ are the 2D discrete Fourier transform and inverse discrete Fourier transform respectively, and $\mathcal{H}$ is an ideal 2D high-pass filter that truncates low-frequency components under a certain threshold. The Fourier transforms are applied on each channel of $z_{i}$ independently. As illustrated in \cref{fig:loss_vis}(e), features associated with structural information can be effectively emphasized by \cref{eq:hpf}. The corresponding features from the predicted latent $D_{\theta}(\hat{n}_{i};\sigma)$ can also be extracted similarly. With the extracted high-frequency features, we devise a new structure preservation loss as:
\begin{equation}
\label{eq:structure}
\mathcal{L}_{\text{structure}}=\mathbb{E}_{\boldsymbol{z},\sigma,\hat{\boldsymbol{n}}}\Big[\sum^{K}_{i=1}(1-m_{i})\odot\Vert\mathcal{F}(D_{\theta}(\hat{n}_{i};\sigma))-\mathcal{F}(z_{i})\Vert^{2}\Big].
\end{equation}
This loss function minimizes the disparity of high-frequency features between prediction and ground truth, so that more structural information can be retained. Our final training objective is a weighted sum of \cref{eq:diffusion}, \cref{eq:dynamics} and \cref{eq:structure}, where $\lambda_{1}$ and $\lambda_{2}$ are trade-off weights to balance the optimization:
\begin{equation}
\label{eq:final}
\mathcal{L}_{\text{final}}=\mathcal{L}_{\text{diffusion}}+\lambda_{1}\mathcal{L}_{\text{dynamics}}+\lambda_{2}\mathcal{L}_{\text{structure}}.
\end{equation}

\subsection{Phase Two: Learning Versatile Action Controllability}
\label{sec:action}
\noindent\textbf{Unified Conditioning of Versatile Actions.}
To maximize usage flexibility, a driving world model should be able to leverage multiple action formats with different characteristics. For instance, one may use the world model to evalute high-level policies~\cite{wang2023driving}, or to execute low-level maneuvers~\cite{santana2016learning}. However, existing approaches only support limited action controls~\cite{hu2023gaia,jia2023adriver,lu2023wovogen,wang2023drivedreamer,wang2023driving}, inhibiting their flexibility and applicability. Therefore, we incorporate a versatile set of action modes for \modelname: \textbf{(1)~Angle and Speed} stand for the utmost fine-grained action controls. We normalize angles to $[-1, 1]$ and represent speeds in $km/h$. The signals from different timestamps are concatenated sequentially. \textbf{(2)~Trajectory} is a series of 2D displacements in ego coordinates. It is widely used as the output of planning algorithms~\cite{casas2021mp3,chitta2022transfuser,hu2023planning,jia2023driveadapter,jia2023think}. We represent the trajectory in meters and flatten it into a sequence. \textbf{(3)~Command} is the most high-level intention. Without loss of generality, we define four commands, \ie go forward, turn right, turn left, and stop, which are implemented as categorical indices. \textbf{(4)~Goal Point} is a 2D coordinate projected from the short-term ego destination onto the initial frame, serving as an interactive interface~\cite{kong2023dreamdrone}. The coordinate is normalized by the image size.

Note that these actions are heterogeneous and cannot be used interchangeably. After transforming all these actions into numerical sequences, we encode them as a unified concatenation of Fourier embeddings~\cite{tancik2020fourier,vaswani2017attention} (see \cref{fig:pipeline}). These embeddings can be jointly ingested by learning additional projections to expand the input dimension of the cross-attention layers in the UNet~\cite{blattmann2023stable}. The new projections are initialized as zeros to enable gradual learning from the pretrained state. We empirically discover that incorporating action conditions through cross-attention layers yields faster convergence and stronger controllability compared to other approaches such as additive embeddings~\cite{wang2023motionctrl,yang2024generalized}.

\noindent\textbf{Efficient Learning.}
We learn action controllability after the first training phase. Since the number of total iterations is crucial for diffusion training~\cite{blattmann2023stable,dai2023emu,girdhar2023emu,podell2023sdxl}, we separate action control learning into two stages. In the first stage, we train our model at a low resolution (320$\times$576), which achieves 3.5$\times$ higher training throughput compared to the original resolution (576$\times$1024). This stage constitutes the majority of training iterations. Then, we finetune the model at the desired resolution (576$\times$1024) for a short duration, so that the learned controllability can effectively cater to high-resolution prediction.

However, tuning the UNet~\cite{blattmann2023stable} at a lower resolution directly may undermine the high-fidelity prediction ability. Conversely, freezing all UNet weights and training the new projections alone would precipitate a quality decline (see \cref{sec:add_exp}), suggesting the necessity to make the UNet adaptable. To solve this, we freeze the pretrained UNet and introduce parameter-efficient LoRA adapters~\cite{hu2021lora} to each attention layer. After training, the low-rank matrices can be seamlessly integrated with the frozen weights, without introducing extra inference latency. Thus, the pretrained weights remain intact when training at the low resolution, avoiding deterioration of the pretrained high-fidelity prediction ability.

Since the parameters of the camera and vehicle are unavailable for open-world scenarios, it seems impossible to obtain multiple equivalent action conditions simultaneously at inference time. Additionally, it will entail prohibitively expensive training to encompass all possible combinations of action conditions. Hence, unlike common practices that activate all conditions during training, we enforce the independence of different action formats by enabling only one of them for each training sample. The remaining action conditions will be filled with zeros as unconditional inputs. As demonstrated in \cref{sec:add_exp}, this simple constraint prevents the squandering of training cost on action combinations and maximizes the learning efficiency of each individual action mode within the same training steps.

\noindent\textbf{Collaborative Training.}
Note that the aforementioned action conditions are not available in OpenDV-YouTube~\cite{yang2024generalized}. On the other hand, nuScenes~\cite{caesar2020nuscenes} has adequate annotations to derive these conditions. To maintain generalization and learn controllability in tandem, we introduce a collaborative training strategy by utilizing the samples from both datasets, with the action conditions for OpenDV-YouTube set to zero. The action control learning phase adopts the same loss as \cref{eq:final}. By learning from two complementary datasets, \modelname gains versatile controllability that are generalizable to novel datasets.

\subsection{Generalizable Reward Function}
\label{sec:reward}
One application of world models is to evaluate actions by engaging a reward module~\cite{hafner2019dream,hafner2020mastering,hafner2023mastering,lecun2022path}. Drive-WM~\cite{wang2023driving} establishes a reward using external detectors~\cite{li2022bevformer,liao2022maptr}. However, these detectors are developed on a particular dataset~\cite{caesar2020nuscenes}, which may become a bottleneck for reward estimation in arbitrary scenarios. On the other hand, \modelname has ingested millions of human driving logs, exhibiting strong generalization across scenes. Based on the observation that out-of-distribution conditions will lead to increased diversity in generation~\cite{escontrela2024video,huang2023diffusion}, we utilize the prediction uncertainty from \modelname itself as the source of our reward. Different from Drive-WM, our reward function seamlessly inherits the generalization of \modelname without resorting to external models. Specifically, we estimate uncertainty via conditional variance. For reliable approximation, we denoise from randomly sampled noise with the same condition frame $\boldsymbol{c}$ and action $\boldsymbol{a}$ for $M$ rounds. Our reward function $R(\boldsymbol{c},\boldsymbol{a})$ is then defined as the exponential of averaged negative conditional variance:
\begin{align}
\label{eq:reward}
\mu'=\frac{1}{M}\sum_{m}&D^{(m)}_{\theta}(\hat{\boldsymbol{n}};\boldsymbol{c},\boldsymbol{a}), \\
R(\boldsymbol{c},\boldsymbol{a})=\texttt{exp}\Big[\texttt{avg}\Big(-\frac{1}{M-1}&\sum_{m}(D^{(m)}_{\theta}(\hat{\boldsymbol{n}};\boldsymbol{c},\boldsymbol{a})-\mu')^{2}\Big)\Big],
\end{align}
where $\texttt{avg}(\cdot)$ averages all latent values within the video clip. Based on this formulation, unfavorable actions with larger uncertainties will lead to lower rewards. In contrast to commonly used evaluation protocols (\eg, the L2 error), our reward function can evaluate actions without referring to the ground truth actions. Note that we do not normalize the estimated rewards for the simplicity of definition, but it is straightforward to amplify the relative contrast by rescaling the estimated rewards with a factor.

\section{Experiments}
\label{sec:experiments}
In this section, we first demonstrate \modelname's strengths in generalization and fidelity in \cref{sec:evaluation}. We then show the impact of action controls in \cref{sec:controllability}. We also substantiate the efficacy of the proposed reward function in \cref{sec:application}. Finally, we conduct ablation studies on our key designs in \cref{sec:ablation}. For more implementation details and experimental results, please refer to \cref{sec:implementation} and \cref{sec:add_exp}.

\begin{table*}[t!]
\vspace{-2mm}
\caption{\textbf{Comparison of prediction fidelity on nuScenes validation set.} 
\modelname achieves encouraging results that outperform the state-of-the-art driving world models with a significant performance gain.}
\label{tab:fvd}
\vspace{-1mm}
\begin{center}
\scalebox{0.75}{
\begin{tabular}{p{0.08\textwidth} | >{\centering}p{0.14\textwidth} >{\centering}p{0.14\textwidth } >{\centering}p{0.14\textwidth} >{\centering}p{0.14\textwidth} >{\centering}p{0.14\textwidth} >{\centering\arraybackslash}p{0.14\textwidth}}
\toprule
\multirow{2}{*}{\textbf{Metric}}
& DriveGAN & DriveDreamer & WoVoGen & Drive-WM & GenAD & \textbf{\modelname} \\
& \cite{santana2016learning} & \cite{wang2023drivedreamer} & \cite{lu2023wovogen} & \cite{wang2023driving} & \cite{yang2024generalized} & \textbf{(Ours)} \\
\midrule
\textbf{FID $\downarrow$} & 73.4 & 52.6 & 27.6 & 15.8 & 15.4 & \textbf{6.9} \\
\textbf{FVD $\downarrow$} & 502.3 & 452.0 & 417.7 & 122.7 & 184.0 & \textbf{89.4} \\
\bottomrule
\end{tabular}
}
\end{center}
\vspace{-4mm}
\end{table*}

\begin{figure}[t!]
\centering
\includegraphics[width=0.99\textwidth]{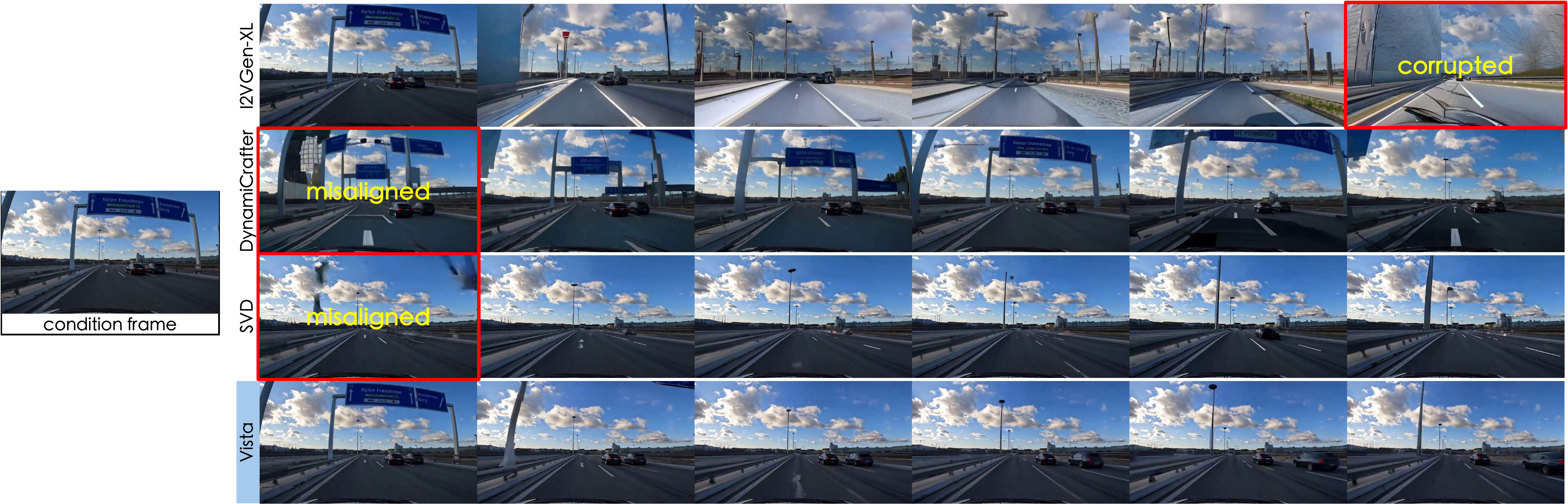}
\vspace{-2mm}
\caption{\textbf{Driving futures predicted by different models using the same condition frame}. We contrast \modelname to publicly available video generation models using their default configurations. Whilst previous models produce misaligned and corrupted results, \modelname does not suffer from these caveats.}
\label{fig:human_vis}
\vspace{-3mm}
\end{figure}

\begin{figure}[t!]
\centering
\includegraphics[width=0.99\textwidth]{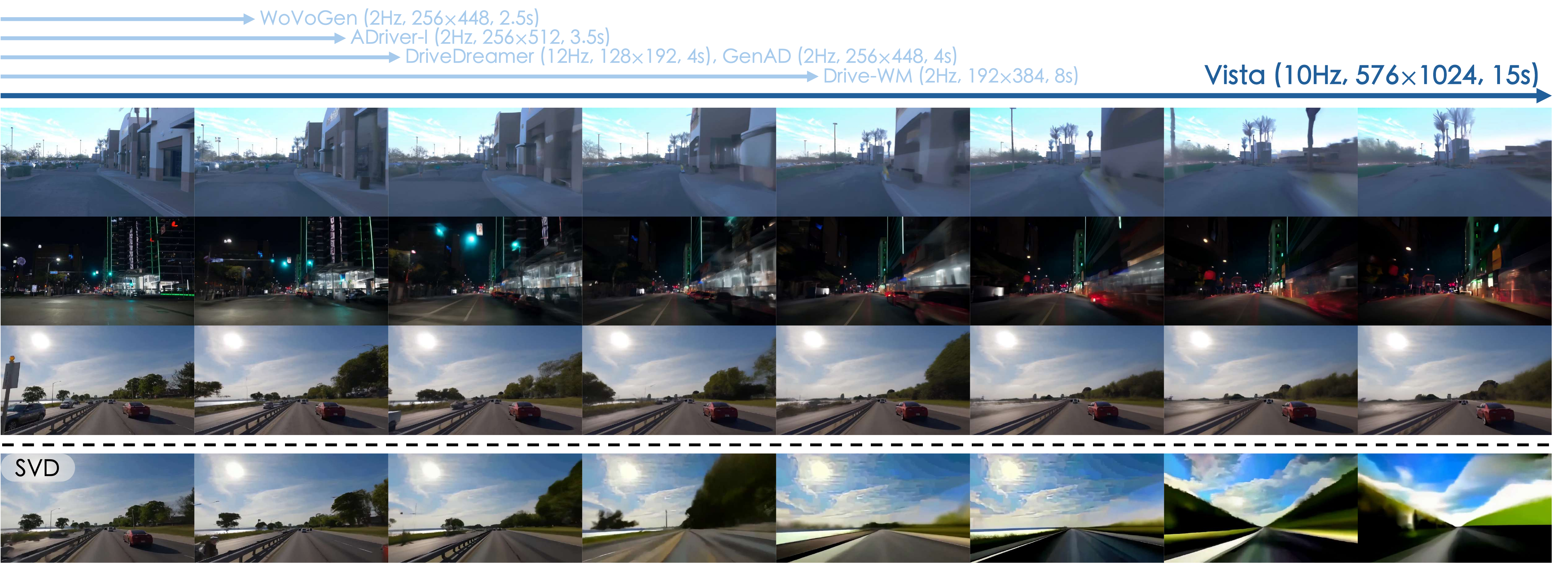}
\vspace{-2mm}
\caption{\textbf{[Top]: Long-horizon prediction.} \modelname can forecast 15 seconds high-resolution futures without much degradation, encompassing long driving distances. The length of the blue lines indicate the duration of the longest prediction showcased by previous works. \textbf{[Bottom]: Long-term extension results of SVD.} SVD fails to generate consistent high-fidelity videos autoregressively as \modelname does.}
\label{fig:long}
\vspace{-2mm}
\end{figure}

\subsection{Comparisons of Generalization and Fidelity}
\label{sec:evaluation}
\noindent\textbf{Automatic Evaluation.}
Since none of the driving world models are publicly accessible, we compare these methods with their quantitative results on nuScenes. We filter 5369 valid samples from the validation set to conduct FID~\cite{heusel2017gans} and FVD~\cite{unterthiner2018towards} evaluation. For FID evaluation, we crop and resize the predicted frames to the resolution of 256$\times$448. For FVD evaluation, we use all 25 frames in each video clip and downsample them to 224$\times$224 following LVDM~\cite{he2022latent}. \Cref{tab:fvd} reports the results of all methods. In both metrics, \modelname surpasses previous driving world models with a considerable margin.

\begin{figure}[t!]
\begin{minipage}[c]{0.585\linewidth}
\centering
\includegraphics[width=0.95\textwidth]{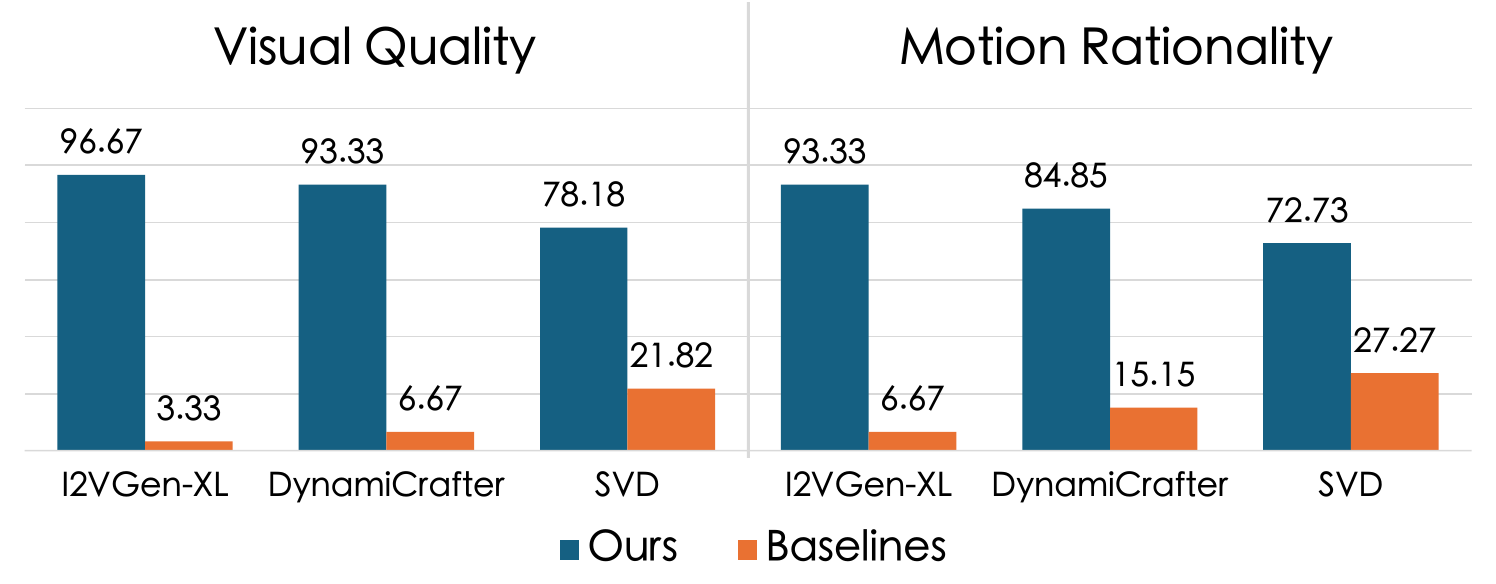}
\vspace{-3mm}
\caption{\textbf{Human evaluation results.} The value denotes the percentage of the times that one model is preferred over the other. \modelname outperforms existing works in both metrics.}
\label{fig:human_stat}
\vspace{-4mm}
\end{minipage}
\hfill
\begin{minipage}[c]{0.4\linewidth}
\centering
\includegraphics[width=0.99\textwidth]{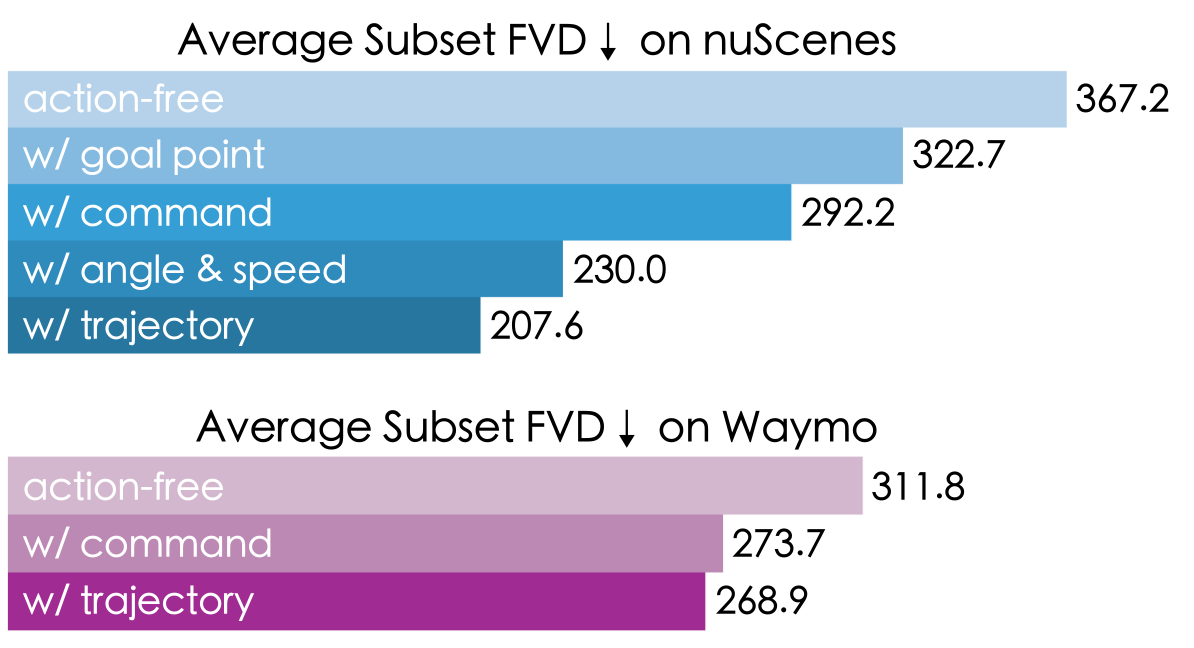}
\vspace{-5.5mm}
\caption{\textbf{Efficacy of action controls.} Applying action controls will produce more similar predictions to the real data.}
\label{fig:action_effect}
\vspace{-3.5mm}
\end{minipage}
\end{figure}

\noindent\textbf{Human Evaluation.}
To analyze the generalization of \modelname across different datasets, we compare it against three prominent general-purpose video generators trained on web-scale data~\cite{blattmann2023stable,xing2023dynamicrafter,zhang2023i2vgen} (see \cref{fig:human_vis}). It is known that automatic metrics like FVD~\cite{unterthiner2018towards} cannot conclusively reveal perceptual quality~\cite{bar2024lumiere,blattmann2023align,girdhar2023emu,zhangjie2024towards,yang2024generalized}, let alone real-world dynamics. Therefore, we opt for human evaluation for more faithful analysis. Following recent advances~\cite{bar2024lumiere,blattmann2023stable,blattmann2023align,chen2023videocrafter1,chen2024videocrafter2,girdhar2023emu,wang2023videofactory,wang2023lavie}, we adopt the Two-Alternative Forced Choice protocol. Specifically, participants are presented with a side-by-side video pair and asked to choose the video they deem better on two orthogonal aspects: visual quality and motion rationality. To avoid potential bias, we crop each video to a fixed aspect ratio, downsample them to the same resolution, and trim the excess frames when \modelname generates longer videos than others. We only feed one condition frame to align with other models. To ensure the variety of scenes, we uniformly assemble 60 scenes from four representative datasets: OpenDV-YouTube-val~\cite{yang2024generalized}, nuScenes~\cite{caesar2020nuscenes}, Waymo~\cite{sun2020scalability}, and CODA~\cite{li2022coda}. These datasets collectively exemplify the intricacy and diversity of real-world driving, \eg, OpenDV-YouTube-val includes geofenced districts, Waymo offers a unique domain compared to our training data, and CODA contains extremely challenging corner cases. We collect a total of 2640 answers from 33 participants. As presented in \cref{fig:human_stat}, \modelname outperforms all baselines on both aspects, demonstrating its profound comprehension of the driving dynamics. Further, unlike other models that are only applicable for short-term generation, \modelname can accommodate more dynamic priors and produce coherent long-horizon rollouts as shown in \cref{fig:long}.

\begin{figure}[t!]
\centering
\includegraphics[width=0.99\textwidth]{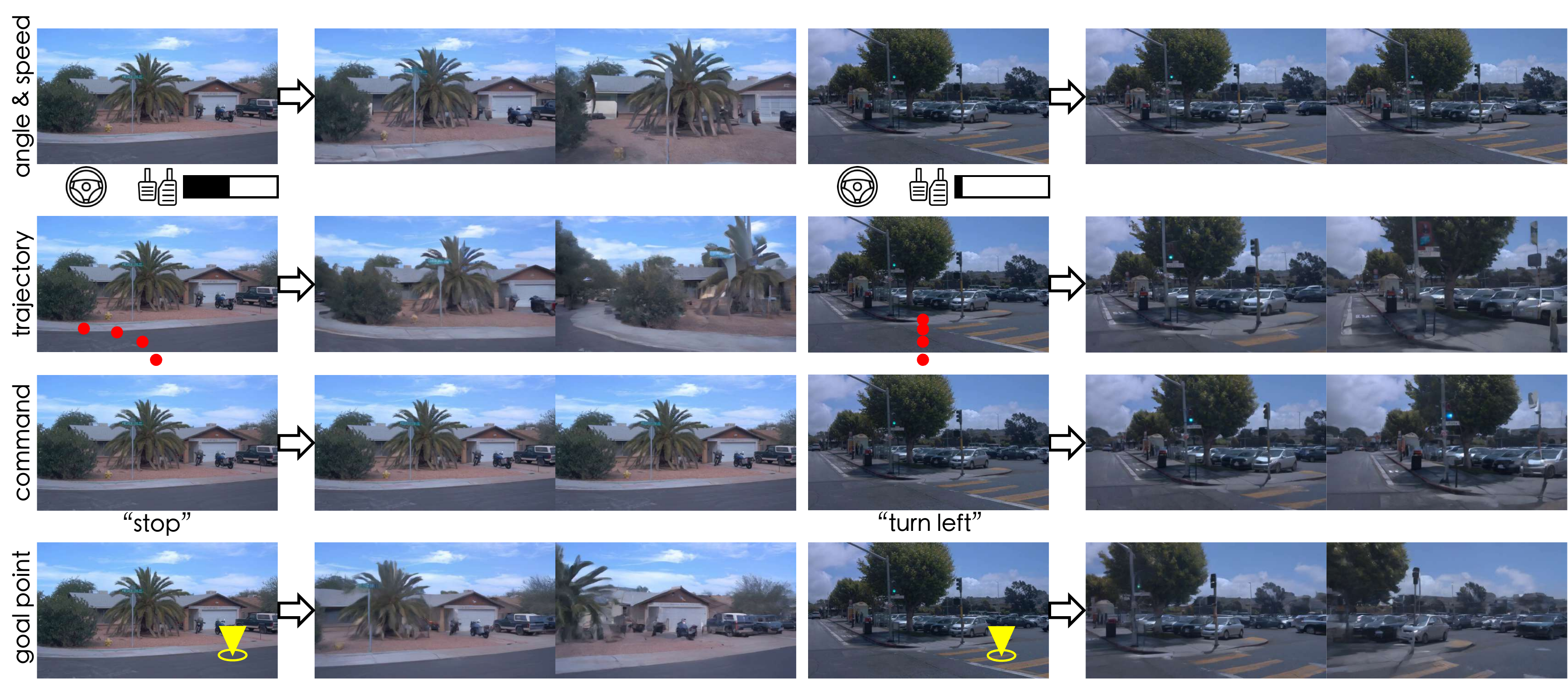}
\vspace{-3mm}
\caption{\textbf{Versatile action controllability.} \modelname can predict the corresponding outcomes in response to multi-modal action conditions across diverse scenarios. More results can be found in \cref{sec:add_vis}.}
\label{fig:control}
\vspace{-3mm}
\end{figure}

\subsection{Results of Action Controllability}
\label{sec:controllability}
\noindent\textbf{Quantitative Results.}
To evaluate the impact of action controls, we divide the validation set of both nuScenes and the unseen Waymo dataset into four subsets according to our command categories. We then generate predictions using different modalities of the ground truth actions. The FVD score is measured on each subset and then averaged. A lower FVD score reflects a closer distribution to the ground truth videos, indicating that the predictions exhibit more resemblance to each specific type of behavior. \cref{fig:action_effect} shows that our action controls can emulate the corresponding movements effectively.

\begin{wrapfigure}{r}{0.33\textwidth}
\centering
\vspace{-5.5mm}
\includegraphics[width=0.3\textwidth]{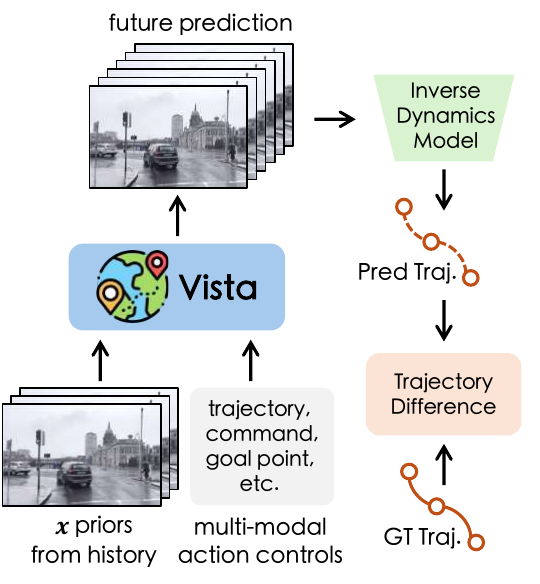}
\vspace{-3.5mm}
\caption{\textbf{An illustration of the IDM experiments in \Cref{tab:rebuttal_idm}.}}
\label{fig:idm}
\vspace{-10mm}
\end{wrapfigure}

We also introduce a new metric named \textit{Trajectory Difference} to assess control consistency. Following GenAD~\cite{yang2024generalized}, we train an inverse dynamics model (IDM) that estimates the corresponding trajectory from a video clip. An illustration of IDM is shown in \cref{fig:idm}. We then send \modelname's prediction to the IDM and calculate the L2 difference between the ground truth trajectory and the estimated trajectory. The differences are measured over 2 seconds. The lower the trajectory difference, the stronger the control consistency \modelname exhibits. We conduct the experiments on nuScenes and Waymo. For each dataset, we collect a subset that contains 537 samples. As reported in \Cref{tab:rebuttal_idm}, \modelname can be effectively controlled by different modalities of actions, resulting in more consistent motions to the ground truth.

\noindent\textbf{Qualitative Results.}
\cref{fig:control} exhibits the versatile action controllability of our model. \modelname can be effectively controlled by multi-modal actions, even in unseen scenarios beyond the training domain. In \cref{sec:add_vis}, we also showcase the counterfactual reasoning ability of \modelname using abnormal actions.

\subsection{Results of Reward Modeling}
\label{sec:application}
\begin{figure}[t!]
\begin{minipage}[c]{0.48\linewidth}
\centering
\includegraphics[width=1.0\textwidth]{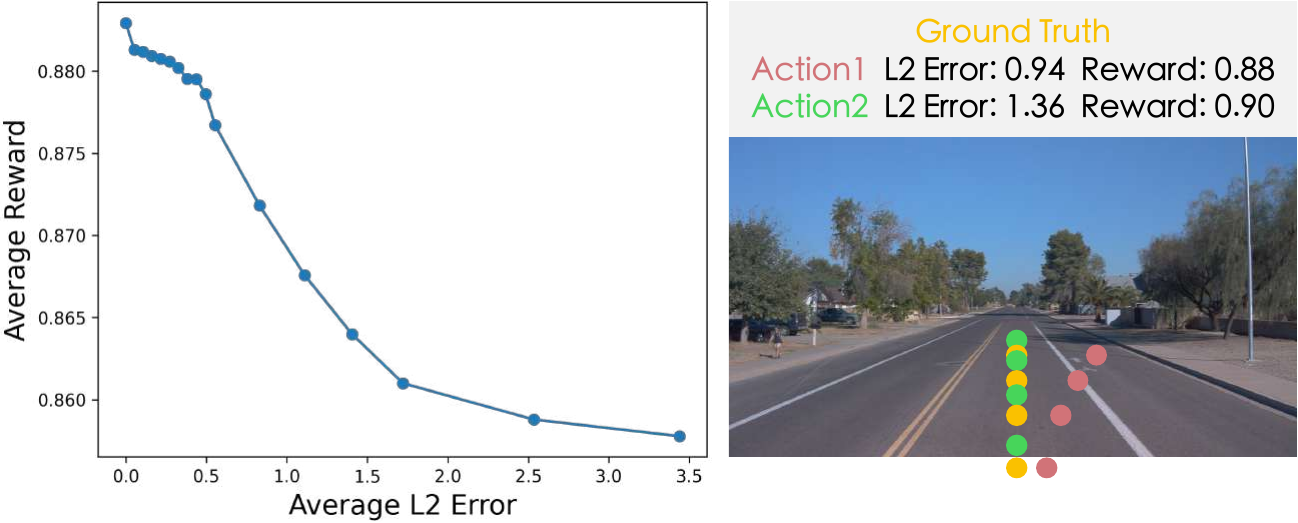}
\vspace{-6mm}
\caption{\textbf{[Left]: Average reward on Waymo with different L2 errors.} \textbf{[Right]: Case study.} The relative contrast of our reward can properly assess the actions that the L2 error fails to judge.}
\label{fig:reward}
\vspace{-3.5mm}
\end{minipage}
\hfill
\begin{minipage}[c]{0.5\linewidth}
\centering
\includegraphics[width=1.0\textwidth]{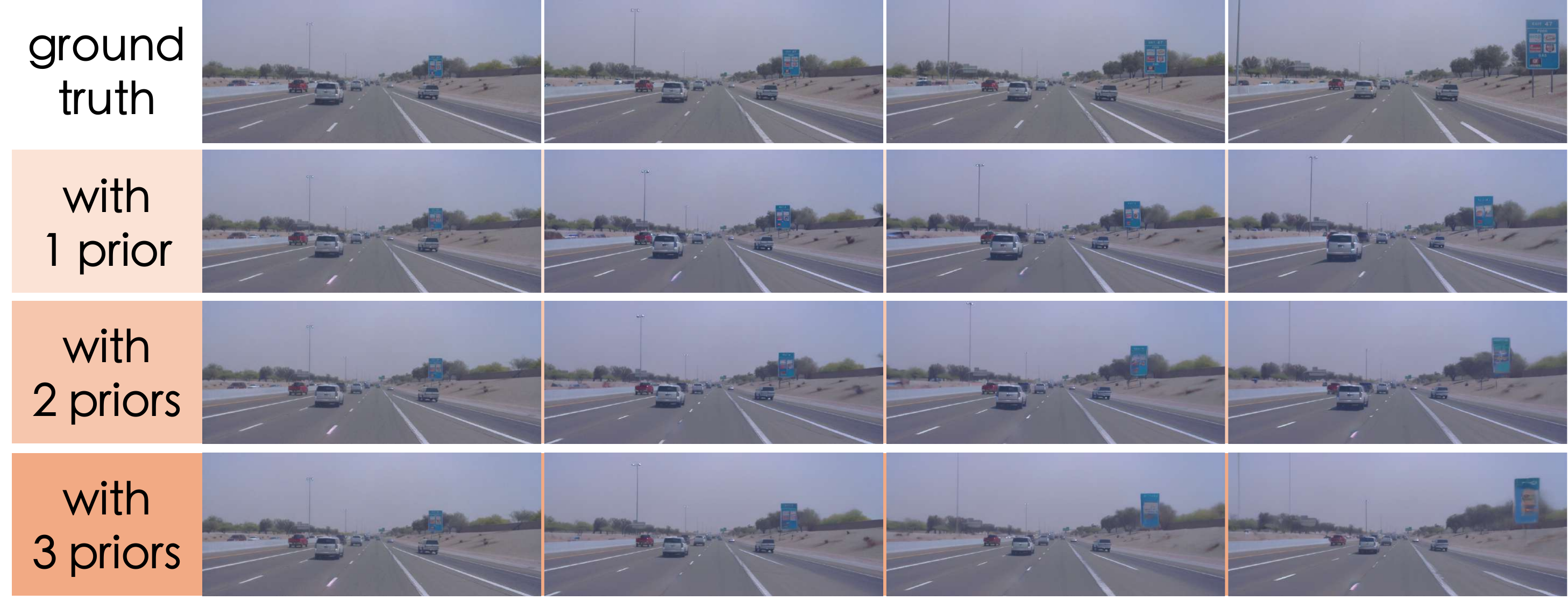}
\vspace{-5.5mm}
\caption{\textbf{Effect of dynamic priors.} Injecting more dynamic priors yields more consistent future motions with the ground truth, such as the motions of the white vehicle and the billboard on the left.}
\label{fig:prior_effect}
\vspace{-2.5mm}
\end{minipage}
\end{figure}

To validate the efficacy of our reward function, we jitter the ground truth trajectories into a series of inferior trajectories. Specifically, we compute the standard deviation of each waypoint from the nuScenes training set as prior distributions. These priors are jointly rescaled to sample perturbations with different L2 errors. The perturbations are then added as offsets to the ground truth trajectories. To ensure the plausibility of sampled trajectories, we adopt an explicit correlating strategy~\cite{gupta2022maskvit,nagabandi2020deep} to regularize offset sampling and recursively sample new trajectories until their offsets are consistent in tendencies. To demonstrate the generality of our reward function, we conduct reward estimation on Waymo~\cite{sun2020scalability}, which is unseen in training. This is done by uniformly sampling from each command category on Waymo validation set, resulting in 1500 cases in total. We compare the average reward of the trajectories with varying L2 errors in \cref{fig:reward}. Our reward decreases when the deviation from the ground truth increases, underscoring the potential of our approach to serve as a viable reward function. It also holds the promise to remedy the irrationality in current evaluation protocols for planning~\cite{chen2023end,li2023ego, zhai2023rethinking}, such as the L2 error shown in \cref{fig:reward}. More in-depth analysis of rewards, including sensitivity to hyperparameters and reward of other actions, are provided in \cref{sec:add_exp}.

\begin{figure}[t!]
\centering
\includegraphics[width=0.99\textwidth]{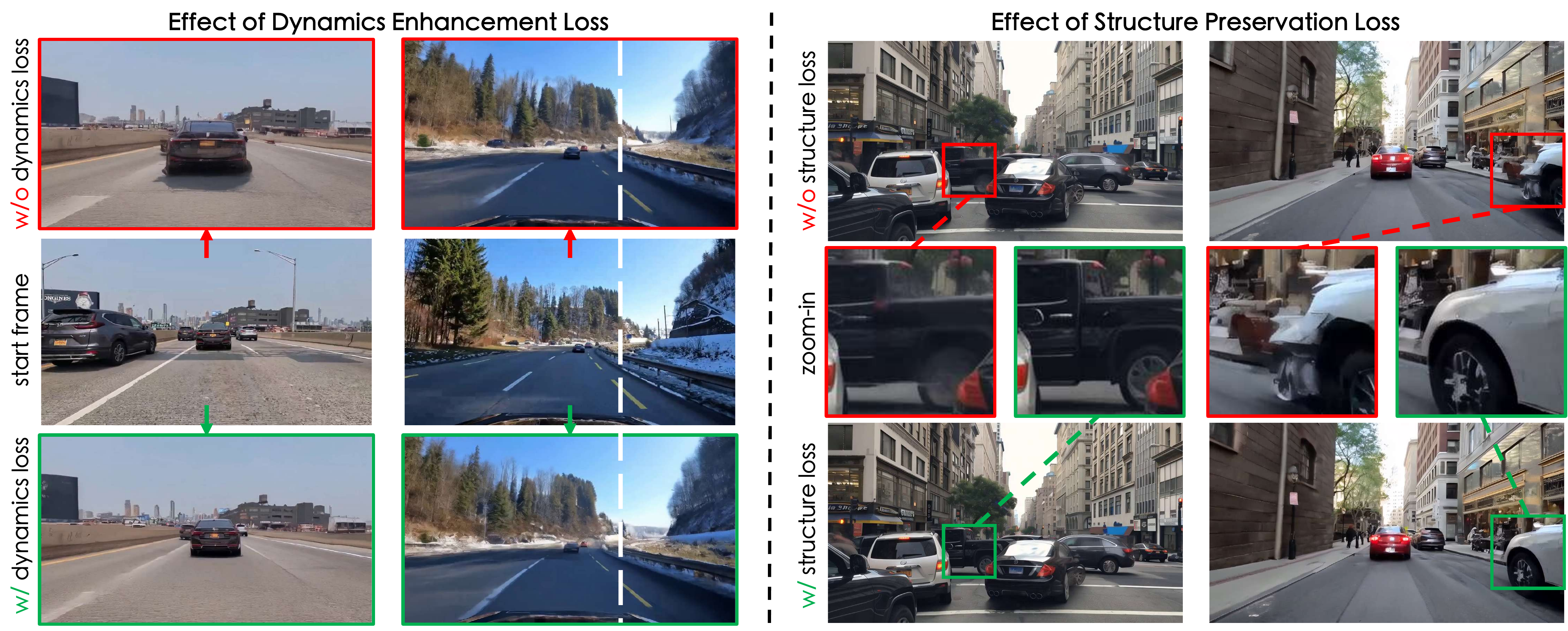}
\vspace{-2.5mm}
\caption{\textbf{[Left]: Effect of dynamics enhancement loss.} The model supervised by the dynamics enhancement loss generates more realistic dynamics. In the first example, instead of remaining static, the front car moves forward normally. In the second example, when the ego-vehicle steers right, the trees shift towards the left naturally adhering to the real-world geometric rules. \textbf{[Right]: Effect of structure preservation loss.} The proposed loss yields a clearer outline of the objects as they move.}
\label{fig:loss_effect}
\vspace{-3mm}
\end{figure}

\subsection{Ablation Study}
\label{sec:ablation}
\noindent\textbf{Dynamic Priors.}
\cref{fig:prior_effect} visualizes the outcomes of using different orders of dynamic priors. The order of priors corresponds to the number of condition frames. It shows that dynamic priors play a pivotal role in long-horizon rollouts, where the coherence with respect to historical frames is essential.

To further demonstrate the efficacy of dynamic priors, we conduct a quantitative evaluation in \Cref{tab:rebuttal_idm}. Specifically, we use the IDM in \cref{sec:controllability} to infer the trajectories of the predicted videos with different orders of dynamic priors. The diminishing differences in trajectory suggest that introducing more priors can effectively improve the consistency between prediction and ground truth.

\noindent\textbf{Auxiliary Supervisions.}
To verify the effectiveness of the two losses proposed in \cref{sec:dyamics}, we devise two additional variants by individually ablating each loss from a variant that incorporates both losses. We qualitatively compare their effects through \cref{fig:loss_effect}, which confirms that the dynamics enhancement loss can promote the learning of real-world dynamics, and the structure preservation loss can reinforce the prediction of structural details.

\begin{table}[t!]
\caption{\textbf{Impacts of different action conditions and dynamic priors.} By applying action conditions and dynamic priors, \modelname can predict motion that is more consistent compared to the ground truth.}
\vspace{-1mm}
\label{tab:rebuttal_idm}
\centering
\scalebox{0.9}{
\begin{tabular}{>{\centering}p{0.14\textwidth} | p{0.20\textwidth} | >{\centering}p{0.14\textwidth} >{\centering}p{0.14\textwidth} >{\centering\arraybackslash}p{0.14\textwidth}}
\toprule
\multirow{2}{*}{\textbf{Dataset}} & \multicolumn{1}{c|}{\multirow{2}{*}{\textbf{Condition}}} & \multicolumn{3}{c}{\textbf{Average Trajectory Difference $\downarrow$}} \\
& & with 1 prior & with 2 priors & with 3 priors \\
\midrule
\multirow{6}{*}{nuScenes} & \textcolor{lightgray}{GT video} & \textcolor{lightgray}{0.379} & \textcolor{lightgray}{0.379} & \textcolor{lightgray}{0.379} \\
& action-free & 3.785 & 2.597 & 1.820 \\
& + goal point & 2.869 & 2.192 & 1.585 \\
& + command & 3.129 & 2.403 & 1.593 \\
& + angle \& speed & 1.562 & 1.123 & 0.832 \\
& + trajectory & 1.559 & 1.148 & 0.835 \\
\midrule
\multirow{4}{*}{Waymo} & \textcolor{lightgray}{GT video} & \textcolor{lightgray}{0.893} & \textcolor{lightgray}{0.893} & \textcolor{lightgray}{0.893} \\
& action-free & 3.646 & 2.901 & 2.052 \\
& + command & 3.160 & 2.561 & 1.902 \\
& + trajectory & 1.187 & 1.147 & 1.140 \\
\bottomrule
\end{tabular}
}
\vspace{-2mm}
\end{table}

\section{Conclusion}
\label{sec:conclusion}
In this paper, we introduce \textit{\modelname}, a generalizable driving world model with enhanced fidelity and controllability. Based on our systematic investigations, \modelname is able to predict realistic and continuous futures at high spatiotemporal resolution. It also possesses versatile action controllability that is generalizable to unseen scenarios. Moreover, it can be formulated as a reward function to evaluate actions. We hope \modelname will usher in broader interest in developing generalizable autonomy systems.

\noindent\textbf{Limitations and future work.}
As an early endeavor, \modelname still exhibits some limitations with respect to computation efficiency, quality maintenance, and training scale. Our future work will look into applying our method to scalable architectures~\cite{hu2023gaia,peebles2023scalable}. More discussions are included in \cref{sec:discuss}.

\section*{Acknowledgments}
This work is supported by National Key R\&D Program of China (2022ZD0160104), National Natural Science Foundation of China (62206172), and Shanghai Committee of Science and Technology (23YF1462000). This work is also partially supported by the BMBF (Tübingen AI Center, FKZ: 01IS18039A), the DFG (SFB 1233, TP 17, project number: 276693517), and the EXC (number 2064/1 – project number: 390727645). We thank the International Max Planck Research School for Intelligent Systems (IMPRS-IS) for supporting Kashyap Chitta. We also appreciate Zetong Yang, Chonghao Sima, Linyan Huang, and the rest members from OpenDriveLab for valuable feedback. We express our sincere gratitude to all anonymous participants for helping with the human evaluation.







\bibliographystyle{plain}
{
\small
\bibliography{bibliography_short, bibliography}

\begin{thebibliography}{100}

\bibitem{ajay2024compositional}
Anurag Ajay, Seungwook Han, Yilun Du, Shuang Li, Abhi Gupta, Tommi Jaakkola, Josh Tenenbaum, Leslie Kaelbling, Akash Srivastava, and Pulkit Agrawal.
\newblock {Compositional Foundation Models for Hierarchical Planning}.
\newblock In {\em NeurIPS}, 2023.

\bibitem{baker2022video}
Bowen Baker, Ilge Akkaya, Peter Zhokov, Joost Huizinga, Jie Tang, Adrien Ecoffet, Brandon Houghton, Raul Sampedro, and Jeff Clune.
\newblock {Video Pretraining (VPT): Learning to Act by Watching Unlabeled Online Videos}.
\newblock In {\em NeurIPS}, 2022.

\bibitem{bar2024lumiere}
Omer Bar-Tal, Hila Chefer, Omer Tov, Charles Herrmann, Roni Paiss, Shiran Zada, Ariel Ephrat, Junhwa Hur, Guanghui Liu, Amit Raj, Yuanzhen Li, Michael Rubinstein, Tomer Michaeli, Oliver Wang, Deqing Sun, Tali Dekel, and Inbar Mosseri.
\newblock {Lumiere: A Space-Time Diffusion Model for Video Generation}.
\newblock {\em arXiv preprint arXiv:2401.12945}, 2024.

\bibitem{black2023zero}
Kevin Black, Mitsuhiko Nakamoto, Pranav Atreya, Homer Walke, Chelsea Finn, Aviral Kumar, and Sergey Levine.
\newblock {Zero-Shot Robotic Manipulation with Pretrained Image-Editing Diffusion Models}.
\newblock In {\em ICLR}, 2024.

\bibitem{blattmann2023stable}
Andreas Blattmann, Tim Dockhorn, Sumith Kulal, Daniel Mendelevitch, Maciej Kilian, Dominik Lorenz, Yam Levi, Zion English, Vikram Voleti, Adam Letts, Varun Jampani, and Robin Rombach.
\newblock {Stable Video Diffusion: Scaling Latent Video Diffusion Models to Large Datasets}.
\newblock {\em arXiv preprint arXiv:2311.15127}, 2023.

\bibitem{blattmann2023align}
Andreas Blattmann, Robin Rombach, Huan Ling, Tim Dockhorn, Seung~Wook Kim, Sanja Fidler, and Karsten Kreis.
\newblock {Align Your Latents: High-Resolution Video Synthesis with Latent Diffusion Models}.
\newblock In {\em CVPR}, 2023.

\bibitem{bogdoll2023muvo}
Daniel Bogdoll, Yitian Yang, and J~Marius Z{\"o}llner.
\newblock {MUVO: A Multimodal Generative World Model for Autonomous Driving with Geometric Representations}.
\newblock {\em arXiv preprint arXiv:2311.11762}, 2023.

\bibitem{brooks2023instructpix2pix}
Tim Brooks, Aleksander Holynski, and Alexei~A Efros.
\newblock {InstructPix2Pix: Learning to Follow Image Editing Instructions}.
\newblock In {\em CVPR}, 2023.

\bibitem{bruce2024genie}
Jake Bruce, Michael Dennis, Ashley Edwards, Jack Parker-Holder, Yuge Shi, Edward Hughes, Matthew Lai, Aditi Mavalankar, Richie Steigerwald, Chris Apps, Yusuf Aytar, Sarah Bechtle, Feryal Behbahani, Stephanie Chan, Nicolas Heess, Lucy Gonzalez, Simon Osindero, Sherjil Ozair, Scott Reed, Jingwei Zhang, Konrad Zolna, Jeff Clune, Nando de~Freitas, Satinder Singh, and Tim Rocktäschel.
\newblock {Genie: Generative Interactive Environments}.
\newblock {\em arXiv preprint arXiv:2402.15391}, 2024.

\bibitem{caesar2020nuscenes}
Holger Caesar, Varun Bankiti, Alex~H. Lang, Sourabh Vora, Venice~Erin Liong, Qiang Xu, Anush Krishnan, Yuxin Pan, Giancarlo Baldan, and Oscar Beijbom.
\newblock {nuScenes: A Multimodal Dataset for Autonomous Driving}.
\newblock In {\em CVPR}, 2020.

\bibitem{caesar2021nuplan}
Holger Caesar, Juraj Kabzan, Kok~Seang Tan, Whye~Kit Fong, Eric Wolff, Alex Lang, Luke Fletcher, Oscar Beijbom, and Sammy Omari.
\newblock {nuPlan: A Closed-Loop ML-based Planning Benchmark for Autonomous Vehicles}.
\newblock In {\em CVPR Workshops}, 2021.

\bibitem{casas2021mp3}
Sergio Casas, Abbas Sadat, and Raquel Urtasun.
\newblock {MP3: A Unified Model to Map, Perceive, Predict and Plan}.
\newblock In {\em CVPR}, 2021.

\bibitem{cen2024using}
Jun Cen, Chenfei Wu, Xiao Liu, Shengming Yin, Yixuan Pei, Jinglong Yang, Qifeng Chen, Nan Duan, and Jianguo Zhang.
\newblock {Using Left and Right Brains Together: Towards Vision and Language Planning}.
\newblock {\em arXiv preprint arXiv:2402.10534}, 2024.

\bibitem{chen2020learning}
Dian Chen, Brady Zhou, Vladlen Koltun, and Philipp Kr{\"a}henb{\"u}hl.
\newblock {Learning by Cheating}.
\newblock In {\em CoRL}, 2019.

\bibitem{chen2023videocrafter1}
Haoxin Chen, Menghan Xia, Yingqing He, Yong Zhang, Xiaodong Cun, Shaoshu Yang, Jinbo Xing, Yaofang Liu, Qifeng Chen, Xintao Wang, Chao Weng, and Ying Shan.
\newblock {VideoCrafter1: Open Diffusion Models for High-Quality Video Generation}.
\newblock {\em arXiv preprint arXiv:2310.19512}, 2023.

\bibitem{chen2024videocrafter2}
Haoxin Chen, Yong Zhang, Xiaodong Cun, Menghan Xia, Xintao Wang, Chao Weng, and Ying Shan.
\newblock {VideoCrafter2: Overcoming Data Limitations for High-Quality Video Diffusion Models}.
\newblock In {\em CVPR}, 2024.

\bibitem{chen2023geodiffusion}
Kai Chen, Enze Xie, Zhe Chen, Yibo Wang, Lanqing Hong, Zhenguo Li, and Dit-Yan Yeung.
\newblock {GeoDiffusion: Text-Prompted Geometric Control for Object Detection Data Generation}.
\newblock In {\em ICLR}, 2024.

\bibitem{chen2023end}
Li~Chen, Penghao Wu, Kashyap Chitta, Bernhard Jaeger, Andreas Geiger, and Hongyang Li.
\newblock {End-to-End Autonomous Driving: Challenges and Frontiers}.
\newblock {\em arXiv preprint arXiv:2306.16927}, 2023.

\bibitem{chen2023gentron}
Shoufa Chen, Mengmeng Xu, Jiawei Ren, Yuren Cong, Sen He, Yanping Xie, Animesh Sinha, Ping Luo, Tao Xiang, and Juan-Manuel Perez-Rua.
\newblock {GenTron: Delving Deep into Diffusion Transformers for Image and Video Generation}.
\newblock In {\em CVPR}, 2024.

\bibitem{chitta2024sledge}
Kashyap Chitta, Daniel Dauner, and Andreas Geiger.
\newblock {SLEDGE: Synthesizing Simulation Environments for Driving Agents with Generative Models}.
\newblock {\em arXiv preprint arXiv:2403.17933}, 2024.

\bibitem{chitta2022transfuser}
Kashyap Chitta, Aditya Prakash, Bernhard Jaeger, Zehao Yu, Katrin Renz, and Andreas Geiger.
\newblock {TransFuser: Imitation with Transformer-based Sensor Fusion for Autonomous Driving}.
\newblock {\em IEEE TPAMI}, 2023.

\bibitem{dai2023emu}
Xiaoliang Dai, Ji~Hou, Chih-Yao Ma, Sam Tsai, Jialiang Wang, Rui Wang, Peizhao Zhang, Simon Vandenhende, Xiaofang Wang, Abhimanyu Dubey, Matthew Yu, Abhishek Kadian, Filip Radenovic, Dhruv Mahajan, Kunpeng Li, Yue Zhao, Vladan Petrovic, Mitesh~Kumar Singh, Simran Motwani, Yi~Wen, Yiwen Song, Roshan Sumbaly, Vignesh Ramanathan, Zijian He, Peter Vajda, and Devi Parikh.
\newblock {Emu: Enhancing Image Generation Models using Photogenic Needles in a Haystack}.
\newblock {\em arXiv preprint arXiv:2309.15807}, 2023.

\bibitem{dauner2024navsim}
Daniel Dauner, Marcel Hallgarten, Tianyu Li, Xinshuo Weng, Zhiyu Huang, Zetong Yang, Hongyang Li, Igor Gilitschenski, Boris Ivanovic, Marco Pavone, et~al.
\newblock {NAVSIM: Data-Driven Non-Reactive Autonomous Vehicle Simulation and Benchmarking}.
\newblock {\em arXiv preprint arXiv:2406.15349}, 2024.

\bibitem{dhariwal2021diffusion}
Prafulla Dhariwal and Alexander Nichol.
\newblock {Diffusion Models Beat GANs on Image Synthesis}.
\newblock In {\em NeurIPS}, 2021.

\bibitem{du2024learning}
Yilun Du, Sherry Yang, Bo~Dai, Hanjun Dai, Ofir Nachum, Josh Tenenbaum, Dale Schuurmans, and Pieter Abbeel.
\newblock {Learning Universal Policies via Text-Guided Video Generation}.
\newblock In {\em NeurIPS}, 2023.

\bibitem{du2023video}
Yilun Du, Sherry Yang, Pete Florence, Fei Xia, Ayzaan Wahid, brian ichter, Pierre Sermanet, Tianhe Yu, Pieter Abbeel, Joshua~B. Tenenbaum, Leslie~Pack Kaelbling, Andy Zeng, and Jonathan Tompson.
\newblock {Video Language Planning}.
\newblock In {\em ICLR}, 2024.

\bibitem{ebert2018visual}
Frederik Ebert, Chelsea Finn, Sudeep Dasari, Annie Xie, Alex Lee, and Sergey Levine.
\newblock {Visual Foresight: Model-based Deep Reinforcement Learning for Vision-based Robotic Control}.
\newblock {\em arXiv preprint arXiv:1812.00568}, 2018.

\bibitem{escontrela2024video}
Alejandro Escontrela, Ademi Adeniji, Wilson Yan, Ajay Jain, Xue~Bin Peng, Ken Goldberg, Youngwoon Lee, Danijar Hafner, and Pieter Abbeel.
\newblock {Video Prediction Models as Rewards for Reinforcement Learning}.
\newblock In {\em NeurIPS}, 2023.

\bibitem{esser2021taming}
Patrick Esser, Robin Rombach, and Bjorn Ommer.
\newblock {Taming Transformers for High-Resolution Image Synthesis}.
\newblock In {\em CVPR}, 2021.

\bibitem{finn2017deep}
Chelsea Finn and Sergey Levine.
\newblock {Deep Visual Foresight for Planning Robot Motion}.
\newblock In {\em ICRA}, 2017.

\bibitem{gao2022enhance}
Zeyu Gao, Yao Mu, Ruoyan Shen, Chen Chen, Yangang Ren, Jianyu Chen, Shengbo~Eben Li, Ping Luo, and Yanfeng Lu.
\newblock {Enhance Sample Efficiency and Robustness of End-to-End Urban Autonomous Driving via Semantic Masked World Model}.
\newblock In {\em NeurIPS Workshops}, 2022.

\bibitem{girdhar2023emu}
Rohit Girdhar, Mannat Singh, Andrew Brown, Quentin Duval, Samaneh Azadi, Sai~Saketh Rambhatla, Akbar Shah, Xi~Yin, Devi Parikh, and Ishan Misra.
\newblock {Emu Video: Factorizing Text-to-Video Generation by Explicit Image Conditioning}.
\newblock {\em arXiv preprint arXiv:2311.10709}, 2023.

\bibitem{goodfellow2014generative}
Ian Goodfellow, Jean Pouget-Abadie, Mehdi Mirza, Bing Xu, David Warde-Farley, Sherjil Ozair, Aaron Courville, and Yoshua Bengio.
\newblock {Generative Adversarial Nets}.
\newblock In {\em NeurIPS}, 2014.

\bibitem{guo2023animatediff}
Yuwei Guo, Ceyuan Yang, Anyi Rao, Yaohui Wang, Yu~Qiao, Dahua Lin, and Bo~Dai.
\newblock {AnimateDiff: Animate Your Personalized Text-to-Image Diffusion Models without Specific Tuning}.
\newblock In {\em ICLR}, 2024.

\bibitem{gupta2022maskvit}
Agrim Gupta, Stephen Tian, Yunzhi Zhang, Jiajun Wu, Roberto Mart{\'\i}n-Mart{\'\i}n, and Li~Fei-Fei.
\newblock {MaskViT: Masked Visual Pre-Training for Video Prediction}.
\newblock In {\em ICLR}, 2023.

\bibitem{gupta2024essential}
Tarun Gupta, Wenbo Gong, Chao Ma, Nick Pawlowski, Agrin Hilmkil, Meyer Scetbon, Ade Famoti, Ashley~Juan Llorens, Jianfeng Gao, Stefan Bauer, Bernhard Kragic, Danica an~Schölkopf, and Cheng Zhang.
\newblock {The Essential Role of Causality in Foundation World Models for Embodied AI}.
\newblock {\em arXiv preprint arXiv:2402.06665}, 2024.

\bibitem{gurnee2023language}
Wes Gurnee and Max Tegmark.
\newblock {Language Models Represent Space and Time}.
\newblock In {\em ICLR}, 2024.

\bibitem{guttenberg2023offset}
Nicholas Guttenberg and CrossLabs.
\newblock {Diffusion with Offset Noise}, 2023.

\bibitem{ha2018recurrent}
David Ha and J{\"u}rgen Schmidhuber.
\newblock {Recurrent World Models Facilitate Policy Evolution}.
\newblock In {\em NeurIPS}, 2018.

\bibitem{hafner2019dream}
Danijar Hafner, Timothy Lillicrap, Jimmy Ba, and Mohammad Norouzi.
\newblock {Dream to Control: Learning Behaviors by Latent Imagination}.
\newblock {\em arXiv preprint arXiv:1912.01603}, 2019.

\bibitem{hafner2019learning}
Danijar Hafner, Timothy Lillicrap, Ian Fischer, Ruben Villegas, David Ha, Honglak Lee, and James Davidson.
\newblock {Learning Latent Dynamics for Planning from Pixels}.
\newblock In {\em ICML}, 2019.

\bibitem{hafner2020mastering}
Danijar Hafner, Timothy Lillicrap, Mohammad Norouzi, and Jimmy Ba.
\newblock {Mastering Atari with Discrete World Models}.
\newblock In {\em ICLR}, 2021.

\bibitem{hafner2023mastering}
Danijar Hafner, Jurgis Pasukonis, Jimmy Ba, and Timothy Lillicrap.
\newblock {Mastering Diverse Domains through World Models}.
\newblock {\em arXiv preprint arXiv:2301.04104}, 2023.

\bibitem{hao2023reasoning}
Shibo Hao, Yi~Gu, Haodi Ma, Joshua~Jiahua Hong, Zhen Wang, Daisy~Zhe Wang, and Zhiting Hu.
\newblock {Reasoning with Language Model is Planning with World Model}.
\newblock In {\em EMNLP}, 2023.

\bibitem{he2024large}
Haoran He, Chenjia Bai, Ling Pan, Weinan Zhang, Bin Zhao, and Xuelong Li.
\newblock {Large-Scale Actionless Video Pre-Training via Discrete Diffusion for Efficient Policy Learning}.
\newblock {\em arXiv preprint arXiv:2402.14407}, 2024.

\bibitem{he2022latent}
Yingqing He, Tianyu Yang, Yong Zhang, Ying Shan, and Qifeng Chen.
\newblock {Latent Video Diffusion Models for High-Fidelity Long Video Generation}.
\newblock {\em arXiv preprint arXiv:2211.13221}, 2022.

\bibitem{heusel2017gans}
Martin Heusel, Hubert Ramsauer, Thomas Unterthiner, Bernhard Nessler, and Sepp Hochreiter.
\newblock {GANs Trained by a Two Time-Scale Update Rule Converge to a Local Nash Equilibrium}.
\newblock In {\em NeurIPS}, 2017.

\bibitem{ho2020denoising}
Jonathan Ho, Ajay Jain, and Pieter Abbeel.
\newblock {Denoising Diffusion Probabilistic Models}.
\newblock In {\em NeurIPS}, 2020.

\bibitem{ho2022cascaded}
Jonathan Ho, Chitwan Saharia, William Chan, David~J Fleet, Mohammad Norouzi, and Tim Salimans.
\newblock {Cascaded Diffusion Models for High Fidelity Image Generation}.
\newblock {\em JMLR}, 2022.

\bibitem{ho2022classifier}
Jonathan Ho and Tim Salimans.
\newblock {Classifier-Free Diffusion Guidance}.
\newblock {\em arXiv preprint arXiv:2207.12598}, 2022.

\bibitem{ho2022video}
Jonathan Ho, Tim Salimans, Alexey Gritsenko, William Chan, Mohammad Norouzi, and David~J Fleet.
\newblock {Video Diffusion Models}.
\newblock {\em arXiv preprint arXiv:2204.03458}, 2022.

\bibitem{hu2022model}
Anthony Hu, Gianluca Corrado, Nicolas Griffiths, Zachary Murez, Corina Gurau, Hudson Yeo, Alex Kendall, Roberto Cipolla, and Jamie Shotton.
\newblock {Model-based Imitation Learning for Urban Driving}.
\newblock In {\em NeurIPS}, 2022.

\bibitem{hu2021fiery}
Anthony Hu, Zak Murez, Nikhil Mohan, Sof{\'\i}a Dudas, Jeffrey Hawke, Vijay Badrinarayanan, Roberto Cipolla, and Alex Kendall.
\newblock {FIERY: Future Instance Prediction in Bird's-Eye View from Surround Monocular Cameras}.
\newblock In {\em ICCV}, 2021.

\bibitem{hu2023gaia}
Anthony Hu, Lloyd Russell, Hudson Yeo, Zak Murez, George Fedoseev, Alex Kendall, Jamie Shotton, and Gianluca Corrado.
\newblock {GAIA-1: A Generative World Model for Autonomous Driving}.
\newblock {\em arXiv preprint arXiv:2309.17080}, 2023.

\bibitem{hu2021lora}
Edward~J Hu, Yelong Shen, Phillip Wallis, Zeyuan Allen-Zhu, Yuanzhi Li, Shean Wang, Lu~Wang, and Weizhu Chen.
\newblock {LoRA: Low-Rank Adaptation of Large Language Models}.
\newblock In {\em ICLR}, 2022.

\bibitem{hu2022st}
Shengchao Hu, Li~Chen, Penghao Wu, Hongyang Li, Junchi Yan, and Dacheng Tao.
\newblock {ST-P3: End-to-end Vision-based Autonomous Driving via Spatial-Temporal Feature Learning}.
\newblock In {\em ECCV}, 2022.

\bibitem{hu2023toward}
Yafei Hu, Quanting Xie, Vidhi Jain, Jonathan Francis, Jay Patrikar, Nikhil Keetha, Seungchan Kim, Yaqi Xie, Tianyi Zhang, Shibo Zhao, Yu~Quan Chong, Chen Wang, Katia Sycara, Matthew Johnson-Roberson, Dhruv Batra, Xiaolong Wang, Sebastian Scherer, Zsolt Kira, Fei Xia, and Yonatan Bisk.
\newblock {Toward General-Purpose Robots via Foundation Models: A Survey and Meta-Analysis}.
\newblock {\em arXiv preprint arXiv:2312.08782}, 2023.

\bibitem{hu2023planning}
Yihan Hu, Jiazhi Yang, Li~Chen, Keyu Li, Chonghao Sima, Xizhou Zhu, Siqi Chai, Senyao Du, Tianwei Lin, Wenhai Wang, Lewei Lu, Xiaosong Jia, Qiang Liu, Jifeng Dai, Yu~Qiao, and Hongyang Li.
\newblock {Planning-Oriented Autonomous Driving}.
\newblock In {\em CVPR}, 2023.

\bibitem{hu2023language}
Zhiting Hu and Tianmin Shu.
\newblock {Language Models, Agent Models, and World Models: The LAW for Machine Reasoning and Planning}.
\newblock {\em arXiv preprint arXiv:2312.05230}, 2023.

\bibitem{huang2023diffusion}
Tao Huang, Guangqi Jiang, Yanjie Ze, and Huazhe Xu.
\newblock {Diffusion Reward: Learning Rewards via Conditional Video Diffusion}.
\newblock {\em arXiv preprint arXiv:2312.14134}, 2023.

\bibitem{jia2023adriver}
Fan Jia, Weixin Mao, Yingfei Liu, Yucheng Zhao, Yuqing Wen, Chi Zhang, Xiangyu Zhang, and Tiancai Wang.
\newblock {ADriver-I: A General World Model for Autonomous Driving}.
\newblock {\em arXiv preprint arXiv:2311.13549}, 2023.

\bibitem{jia2023driveadapter}
Xiaosong Jia, Yulu Gao, Li~Chen, Junchi Yan, Patrick~Langechuan Liu, and Hongyang Li.
\newblock {DriveAdapter: Breaking the Coupling Barrier of Perception and Planning in End-to-End Autonomous Driving}.
\newblock In {\em ICCV}, 2023.

\bibitem{jia2023think}
Xiaosong Jia, Penghao Wu, Li~Chen, Jiangwei Xie, Conghui He, Junchi Yan, and Hongyang Li.
\newblock {Think Twice Before Driving: Towards Scalable Decoders for End-to-End Autonomous Driving}.
\newblock In {\em CVPR}, 2023.

\bibitem{jiang2023vad}
Bo~Jiang, Shaoyu Chen, Qing Xu, Bencheng Liao, Jiajie Chen, Helong Zhou, Qian Zhang, Wenyu Liu, Chang Huang, and Xinggang Wang.
\newblock {VAD: Vectorized Scene Representation for Efficient Autonomous Driving}.
\newblock In {\em ICCV}, 2023.

\bibitem{jordan1992forward}
Michael~I Jordan and David~E Rumelhart.
\newblock {Forward Models: Supervised Learning with a Distal Teacher}.
\newblock {\em Cognitive Science}, 1992.

\bibitem{karras2022elucidating}
Tero Karras, Miika Aittala, Timo Aila, and Samuli Laine.
\newblock {Elucidating the Design Space of Diffusion-based Generative Models}.
\newblock In {\em NeurIPS}, 2022.

\bibitem{khurana2023point}
Tarasha Khurana, Peiyun Hu, David Held, and Deva Ramanan.
\newblock {Point Cloud Forecasting as a Proxy for 4D Occupancy Forecasting}.
\newblock In {\em CVPR}, 2023.

\bibitem{kim2021drivegan}
Seung~Wook Kim, Jonah Philion, Antonio Torralba, and Sanja Fidler.
\newblock {DriveGAN: Towards a Controllable High-Quality Neural Simulation}.
\newblock In {\em CVPR}, 2021.

\bibitem{kim2020learning}
Seung~Wook Kim, Yuhao Zhou, Jonah Philion, Antonio Torralba, and Sanja Fidler.
\newblock {Learning to Simulate Dynamic Environments with GameGAN}.
\newblock In {\em CVPR}, 2020.

\bibitem{kingma2013auto}
Diederik~P Kingma and Max Welling.
\newblock {Auto-Encoding Variational Bayes}.
\newblock {\em arXiv preprint arXiv:1312.6114}, 2013.

\bibitem{ko2023learning}
Po-Chen Ko, Jiayuan Mao, Yilun Du, Shao-Hua Sun, and Joshua~B Tenenbaum.
\newblock {Learning to Act from Actionless Videos through Dense Correspondences}.
\newblock In {\em ICLR}, 2024.

\bibitem{koh2021pathdreamer}
Jing~Yu Koh, Honglak Lee, Yinfei Yang, Jason Baldridge, and Peter Anderson.
\newblock {Pathdreamer: A World Model for Indoor Navigation}.
\newblock In {\em ICCV}, 2021.

\bibitem{kondratyuk2023videopoet}
Dan Kondratyuk, Lijun Yu, Xiuye Gu, José Lezama, Jonathan Huang, Rachel Hornung, Hartwig Adam, Hassan Akbari, Yair Alon, Vighnesh Birodkar, Yong Cheng, Ming-Chang Chiu, Josh Dillon, Irfan Essa, Agrim Gupta, Meera Hahn, Anja Hauth, David Hendon, Alonso Martinez, David Minnen, David Ross, Grant Schindler, Mikhail Sirotenko, Kihyuk Sohn, Krishna Somandepalli, Huisheng Wang, Jimmy Yan, Ming-Hsuan Yang, Xuan Yang, Bryan Seybold, and Lu~Jiang.
\newblock {VideoPoet: A Large Language Model for Zero-Shot Video Generation}.
\newblock {\em arXiv preprint arXiv:2312.14125}, 2023.

\bibitem{kong2023dreamdrone}
Hanyang Kong, Dongze Lian, Michael~Bi Mi, and Xinchao Wang.
\newblock {DreamDrone}.
\newblock {\em arXiv preprint arXiv:2312.08746}, 2023.

\bibitem{lai2023xvo}
Lei Lai, Zhongkai Shangguan, Jimuyang Zhang, and Eshed Ohn-Bar.
\newblock {XVO: Generalized Visual Odometry via Cross-Modal Self-Training}.
\newblock In {\em ICCV}, 2023.

\bibitem{lecun2022path}
Yann LeCun.
\newblock {A Path towards Autonomous Machine Intelligence}.
\newblock {\em Open Review}, 62, 2022.

\bibitem{li2023open}
Hongyang Li, Yang Li, Huijie Wang, Jia Zeng, Huilin Xu, Pinlong Cai, Li~Chen, Junchi Yan, Feng Xu, Lu~Xiong, Jingdong Wang, Futang Zhu, Chunjing Xu, Tiancai Wang, Fei Xia, Beipeng Mu, Zhihui Peng, Dahua Lin, and Yu~Qiao.
\newblock {Open-sourced Data Ecosystem in Autonomous Driving: the Present and Future}.
\newblock {\em arXiv preprint arXiv:2312.03408}, 2023.

\bibitem{li2023delving}
Hongyang Li, Chonghao Sima, Jifeng Dai, Wenhai Wang, Lewei Lu, Huijie Wang, Jia Zeng, Zhiqi Li, Jiazhi Yang, Hanming Deng, Hao Tian, Enze Xie, Jiangwei Xie, Li~Chen, Tianyu Li, Yang Li, Yulu Gao, Xiaosong Jia, Si~Liu, Jianping Shi, Dahua Lin, and Yu~Qiao.
\newblock {Delving Into the Devils of Bird’s-Eye-View Perception: A Review, Evaluation and Recipe}.
\newblock {\em IEEE TPAMI}, 2023.

\bibitem{li2022coda}
Kaican Li, Kai Chen, Haoyu Wang, Lanqing Hong, Chaoqiang Ye, Jianhua Han, Yukuai Chen, Wei Zhang, Chunjing Xu, Dit-Yan Yeung, Xiaodan Liang, Zhenguo Li, and Hang Xu.
\newblock {CODA: A Real-World Road Corner Case Dataset for Object Detection in Autonomous Driving}.
\newblock In {\em ECCV}, 2022.

\bibitem{li2022emergent}
Kenneth Li, Aspen~K Hopkins, David Bau, Fernanda Vi{\'e}gas, Hanspeter Pfister, and Martin Wattenberg.
\newblock {Emergent World Representations: Exploring a Sequence Model Trained on a Synthetic Task}.
\newblock In {\em ICLR}, 2023.

\bibitem{li2024think2drive}
Qifeng Li, Xiaosong Jia, Shaobo Wang, and Junchi Yan.
\newblock {Think2Drive: Efficient Reinforcement Learning by Thinking in Latent World Model for Quasi-Realistic Autonomous Driving (in CARLA-v2)}.
\newblock {\em arXiv preprint arXiv:2402.16720}, 2024.

\bibitem{li2022bevformer}
Zhiqi Li, Wenhai Wang, Hongyang Li, Enze Xie, Chonghao Sima, Tong Lu, Yu~Qiao, and Jifeng Dai.
\newblock {BEVFormer: Learning Bird’s-Eye-View Representation from Multi-Camera Images via Spatiotemporal Transformers}.
\newblock In {\em ECCV}, 2022.

\bibitem{li2023ego}
Zhiqi Li, Zhiding Yu, Shiyi Lan, Jiahan Li, Jan Kautz, Tong Lu, and Jose~M Alvarez.
\newblock {Is Ego Status All You Need for Open-Loop End-to-End Autonomous Driving?}
\newblock In {\em CVPR}, 2024.

\bibitem{liao2022maptr}
Bencheng Liao, Shaoyu Chen, Xinggang Wang, Tianheng Cheng, Qian Zhang, Wenyu Liu, and Chang Huang.
\newblock {MapTR: Structured Modeling and Learning for Online Vectorized HD Map Construction}.
\newblock In {\em ICLR}, 2023.

\bibitem{lin2023learning}
Jessy Lin, Yuqing Du, Olivia Watkins, Danijar Hafner, Pieter Abbeel, Dan Klein, and Anca Dragan.
\newblock {Learning to Model the World with Language}.
\newblock {\em arXiv preprint arXiv:2308.01399}, 2023.

\bibitem{liu2024world}
Hao Liu, Wilson Yan, Matei Zaharia, and Pieter Abbeel.
\newblock {World Model on Million-Length Video and Language With RingAttention}.
\newblock {\em arXiv preprint arXiv:2402.08268}, 2024.

\bibitem{loshchilov2017decoupled}
Ilya Loshchilov and Frank Hutter.
\newblock {Decoupled Weight Decay Regularization}.
\newblock {\em arXiv preprint arXiv:1711.05101}, 2017.

\bibitem{lotter2016deep}
William Lotter, Gabriel Kreiman, and David Cox.
\newblock {Deep Predictive Coding Networks for Video Prediction and Unsupervised Learning}.
\newblock In {\em ICLR}, 2017.

\bibitem{lu2022dpm}
Cheng Lu, Yuhao Zhou, Fan Bao, Jianfei Chen, Chongxuan Li, and Jun Zhu.
\newblock {DPM-Solver: A Fast ODE Solver for Diffusion Probabilistic Model Sampling in Around 10 Steps}.
\newblock In {\em NeurIPS}, 2022.

\bibitem{lu2023wovogen}
Jiachen Lu, Ze~Huang, Jiahui Zhang, Zeyu Yang, and Li~Zhang.
\newblock {WoVoGen: World Volume-Aware Diffusion for Controllable Multi-Camera Driving Scene Generation}.
\newblock {\em arXiv preprint arXiv:2312.02934}, 2023.

\bibitem{luo2023latent}
Simian Luo, Yiqin Tan, Longbo Huang, Jian Li, and Hang Zhao.
\newblock {Latent Consistency Models: Synthesizing High-Resolution Images with Few-Step Inference}.
\newblock {\em arXiv preprint arXiv:2310.04378}, 2023.

\bibitem{mendonca2023structured}
Russell Mendonca, Shikhar Bahl, and Deepak Pathak.
\newblock {Structured World Models from Human Videos}.
\newblock In {\em RSS}, 2023.

\bibitem{meng2023distillation}
Chenlin Meng, Robin Rombach, Ruiqi Gao, Diederik Kingma, Stefano Ermon, Jonathan Ho, and Tim Salimans.
\newblock {On Distillation of Guided Diffusion Models}.
\newblock In {\em CVPR}, 2023.

\bibitem{micheli2022transformers}
Vincent Micheli, Eloi Alonso, and Fran{\c{c}}ois Fleuret.
\newblock {Transformers are Sample-Efficient World Models}.
\newblock In {\em ICLR}, 2023.

\bibitem{nagabandi2020deep}
Anusha Nagabandi, Kurt Konolige, Sergey Levine, and Vikash Kumar.
\newblock {Deep Dynamics Models for Learning Dexterous Manipulation}.
\newblock In {\em CoRL}, 2020.

\bibitem{oh2015action}
Junhyuk Oh, Xiaoxiao Guo, Honglak Lee, Richard~L Lewis, and Satinder Singh.
\newblock {Action-Conditional Video Prediction using Deep Networks in Atari Games}.
\newblock In {\em NeurIPS}, 2015.

\bibitem{peebles2023scalable}
William Peebles and Saining Xie.
\newblock {Scalable Diffusion Models with Transformers}.
\newblock In {\em ICCV}, 2023.

\bibitem{piergiovanni2019learning}
AJ~Piergiovanni, Alan Wu, and Michael~S Ryoo.
\newblock {Learning Real-World Robot Policies by Dreaming}.
\newblock In {\em IROS}, 2019.

\bibitem{podell2023sdxl}
Dustin Podell, Zion English, Kyle Lacey, Andreas Blattmann, Tim Dockhorn, Jonas M{\"u}ller, Joe Penna, and Robin Rombach.
\newblock {SDXL: Improving Latent Diffusion Models for High-Resolution Image Synthesis}.
\newblock In {\em ICLR}, 2024.

\bibitem{rombach2022high}
Robin Rombach, Andreas Blattmann, Dominik Lorenz, Patrick Esser, and Bj{\"o}rn Ommer.
\newblock {High-Resolution Image Synthesis with Latent Diffusion Models}.
\newblock In {\em CVPR}, 2022.

\bibitem{salimans2022progressive}
Tim Salimans and Jonathan Ho.
\newblock {Progressive Distillation for Fast Sampling of Diffusion Models}.
\newblock In {\em ICLR}, 2023.

\bibitem{santana2016learning}
Eder Santana and George Hotz.
\newblock {Learning a Driving Simulator}.
\newblock {\em arXiv preprint arXiv:1608.01230}, 2016.

\bibitem{sauer2024fast}
Axel Sauer, Frederic Boesel, Tim Dockhorn, Andreas Blattmann, Patrick Esser, and Robin Rombach.
\newblock {Fast High-Resolution Image Synthesis with Latent Adversarial Diffusion Distillation}.
\newblock {\em arXiv preprint arXiv:2403.12015}, 2024.

\bibitem{schubert2023generalist}
Ingmar Schubert, Jingwei Zhang, Jake Bruce, Sarah Bechtle, Emilio Parisotto, Martin Riedmiller, Jost~Tobias Springenberg, Arunkumar Byravan, Leonard Hasenclever, and Nicolas Heess.
\newblock {A Generalist Dynamics Model for Control}.
\newblock {\em arXiv preprint arXiv:2305.10912}, 2023.

\bibitem{schwarzer2020data}
Max Schwarzer, Ankesh Anand, Rishab Goel, R~Devon Hjelm, Aaron Courville, and Philip Bachman.
\newblock {Data-Efficient Reinforcement Learning with Self-Predictive Representations}.
\newblock In {\em ICLR}, 2021.

\bibitem{shah2023vint}
Dhruv Shah, Ajay Sridhar, Nitish Dashora, Kyle Stachowicz, Kevin Black, Noriaki Hirose, and Sergey Levine.
\newblock {ViNT: A Foundation Model for Visual Navigation}.
\newblock In {\em CoRL}, 2023.

\bibitem{sima2023drivelm}
Chonghao Sima, Katrin Renz, Kashyap Chitta, Li~Chen, Hanxue Zhang, Chengen Xie, Ping Luo, Andreas Geiger, and Hongyang Li.
\newblock {DriveLM: Driving with Graph Visual Question Answering}.
\newblock {\em arXiv preprint arXiv:2312.14150}, 2023.

\bibitem{singer2022make}
Uriel Singer, Adam Polyak, Thomas Hayes, Xi~Yin, Jie An, Songyang Zhang, Qiyuan Hu, Harry Yang, Oron Ashual, Oran Gafni, Devi Parikh, Sonal Gupta, and Yaniv Taigman.
\newblock {Make-A-Video: Text-to-Video Generation without Text-Video Data}.
\newblock In {\em ICLR}, 2023.

\bibitem{song2020denoising}
Jiaming Song, Chenlin Meng, and Stefano Ermon.
\newblock {Denoising Diffusion Implicit Models}.
\newblock In {\em ICLR}, 2021.

\bibitem{song2023consistency}
Yang Song, Prafulla Dhariwal, Mark Chen, and Ilya Sutskever.
\newblock {Consistency Models}.
\newblock In {\em ICML}, 2023.

\bibitem{song2020score}
Yang Song, Jascha Sohl-Dickstein, Diederik~P Kingma, Abhishek Kumar, Stefano Ermon, and Ben Poole.
\newblock {Score-based Generative Modeling through Stochastic Differential Equations}.
\newblock In {\em ICLR}, 2021.

\bibitem{sun2020scalability}
Pei Sun, Henrik Kretzschmar, Xerxes Dotiwalla, Aurelien Chouard, Vijaysai Patnaik, Paul Tsui, James Guo, Yin Zhou, Yuning Chai, Benjamin Caine, Vijay Vasudevan, Wei Han, Jiquan Ngiam, Hang Zhao, Aleksei Timofeev, Scott~M. Ettinger, Maxim Krivokon, Amy Gao, Aditya Joshi, Yu~Zhang, Jonathon Shlens, Zhifeng Chen, and Dragomir Anguelov.
\newblock {Scalability in Perception for Autonomous Driving: Waymo Open Dataset}.
\newblock In {\em CVPR}, 2020.

\bibitem{sutton2022quest}
Richard~S Sutton.
\newblock {The Quest for a Common Model of the Intelligent Decision Maker}.
\newblock {\em arXiv preprint arXiv:2202.13252}, 2022.

\bibitem{tancik2020fourier}
Matthew Tancik, Pratul Srinivasan, Ben Mildenhall, Sara Fridovich-Keil, Nithin Raghavan, Utkarsh Singhal, Ravi Ramamoorthi, Jonathan Barron, and Ren Ng.
\newblock {Fourier Features Let Networks Learn High Frequency Functions in Low Dimensional Domains}.
\newblock In {\em NeurIPS}, 2020.

\bibitem{unterthiner2018towards}
Thomas Unterthiner, Sjoerd Van~Steenkiste, Karol Kurach, Raphael Marinier, Marcin Michalski, and Sylvain Gelly.
\newblock {Towards Accurate Generative Models of Video: A New Metric \& Challenges}.
\newblock {\em arXiv preprint arXiv:1812.01717}, 2018.

\bibitem{vaswani2017attention}
Ashish Vaswani, Noam Shazeer, Niki Parmar, Jakob Uszkoreit, Llion Jones, Aidan~N Gomez, {\L}ukasz Kaiser, and Illia Polosukhin.
\newblock {Attention is All You Need}.
\newblock In {\em NeurIPS}, 2017.

\bibitem{voleti2022mcvd}
Vikram Voleti, Alexia Jolicoeur-Martineau, and Chris Pal.
\newblock {MCVD: Masked Conditional Video Diffusion for Prediction, Generation, and Interpolation}.
\newblock In {\em NeurIPS}, 2022.

\bibitem{voleti2024sv3d}
Vikram Voleti, Chun-Han Yao, Mark Boss, Adam Letts, David Pankratz, Dmitry Tochilkin, Christian Laforte, Robin Rombach, and Varun Jampani.
\newblock {SV3D: Novel Multi-View Synthesis and 3D Generation from a Single Image using Latent Video Diffusion}.
\newblock {\em arXiv preprint arXiv:2403.12008}, 2024.

\bibitem{vondrick2016generating}
Carl Vondrick, Hamed Pirsiavash, and Antonio Torralba.
\newblock {Generating Videos with Scene Dynamics}.
\newblock In {\em NeurIPS}, 2016.

\bibitem{wang2023dreamwalker}
Hanqing Wang, Wei Liang, Luc Van~Gool, and Wenguan Wang.
\newblock {DREAMWALKER: Mental Planning for Continuous Vision-Language Navigation}.
\newblock In {\em ICCV}, 2023.

\bibitem{wang2023modelscope}
Jiuniu Wang, Hangjie Yuan, Dayou Chen, Yingya Zhang, Xiang Wang, and Shiwei Zhang.
\newblock {ModelScope Text-to-Video Technical Report}.
\newblock {\em arXiv preprint arXiv:2308.06571}, 2023.

\bibitem{wang2023videofactory}
Wenjing Wang, Huan Yang, Zixi Tuo, Huiguo He, Junchen Zhu, Jianlong Fu, and Jiaying Liu.
\newblock {VideoFactory: Swap Attention in Spatiotemporal Diffusions for Text-to-Video Generation}.
\newblock {\em arXiv preprint arXiv:2305.10874}, 2023.

\bibitem{wang2023recipe}
Xiang Wang, Shiwei Zhang, Hangjie Yuan, Zhiwu Qing, Biao Gong, Yingya Zhang, Yujun Shen, Changxin Gao, and Nong Sang.
\newblock {A Recipe for Scaling up Text-to-Video Generation with Text-Free Videos}.
\newblock In {\em CVPR}, 2024.

\bibitem{wang2023videolcm}
Xiang Wang, Shiwei Zhang, Han Zhang, Yu~Liu, Yingya Zhang, Changxin Gao, and Nong Sang.
\newblock {VideoLCM: Video Latent Consistency Model}.
\newblock {\em arXiv preprint arXiv:2312.09109}, 2023.

\bibitem{wang2023drivedreamer}
Xiaofeng Wang, Zheng Zhu, Guan Huang, Xinze Chen, and Jiwen Lu.
\newblock {DriveDreamer: Towards Real-World-Driven World Models for Autonomous Driving}.
\newblock {\em arXiv preprint arXiv:2309.09777}, 2023.

\bibitem{wang2023lavie}
Yaohui Wang, Xinyuan Chen, Xin Ma, Shangchen Zhou, Ziqi Huang, Yi~Wang, Ceyuan Yang, Yinan He, Jiashuo Yu, Peiqing Yang, Yuwei Guo, Tianxing Wu, Chenyang Si, Yuming Jiang, Cunjian Chen, Chen~Change Loy, Bo~Dai, Dahua Lin, Yu~Qiao, and Ziwei Liu.
\newblock {LAVIE: High-Quality Video Generation with Cascaded Latent Diffusion Models}.
\newblock {\em arXiv preprint arXiv:2309.15103}, 2023.

\bibitem{wang2023driving}
Yuqi Wang, Jiawei He, Lue Fan, Hongxin Li, Yuntao Chen, and Zhaoxiang Zhang.
\newblock {Driving into the Future: Multiview Visual Forecasting and Planning with World Model for Autonomous Driving}.
\newblock In {\em CVPR}, 2024.

\bibitem{wang2023motionctrl}
Zhouxia Wang, Ziyang Yuan, Xintao Wang, Tianshui Chen, Menghan Xia, Ping Luo, and Ying Shan.
\newblock {MotionCtrl: A Unified and Flexible Motion Controller for Video Generation}.
\newblock {\em arXiv preprint arXiv:2312.03641}, 2023.

\bibitem{wilson2023argoverse}
Benjamin Wilson, William Qi, Tanmay Agarwal, John Lambert, Jagjeet Singh, Siddhesh Khandelwal, Bowen Pan, Ratnesh Kumar, Andrew Hartnett, Jhony~Kaesemodel Pontes, et~al.
\newblock {Argoverse 2: Next Generation Datasets for Self-Driving Perception and Forecasting}.
\newblock In {\em NeurIPS Datasets and Benchmarks}, 2023.

\bibitem{zhangjie2024towards}
Jay~Zhangjie Wu, Guian Fang, Haoning Wu, Xintao Wang, Yixiao Ge, Xiaodong Cun, David~Junhao Zhang, Jia-Wei Liu, Yuchao Gu, Rui Zhao, Weisi Lin, Wynne Hsu, Ying Shan, and Mike~Zheng Shou.
\newblock {Towards A Better Metric for Text-to-Video Generation}.
\newblock {\em arXiv preprint arXiv:2401.07781}, 2024.

\bibitem{wu2024pre}
Jialong Wu, Haoyu Ma, Chaoyi Deng, and Mingsheng Long.
\newblock {Pre-Training Contextualized World Models with In-the-Wild Videos for Reinforcement Learning}.
\newblock In {\em NeurIPS}, 2023.

\bibitem{wu2023policy}
Penghao Wu, Li~Chen, Hongyang Li, Xiaosong Jia, Junchi Yan, and Yu~Qiao.
\newblock {Policy Pre-training for Autonomous Driving via Self-supervised Geometric Modeling}.
\newblock In {\em ICLR}, 2023.

\bibitem{xing2023dynamicrafter}
Jinbo Xing, Menghan Xia, Yong Zhang, Haoxin Chen, Xintao Wang, Tien-Tsin Wong, and Ying Shan.
\newblock {DynamiCrafter: Animating Open-Domain Images with Video Diffusion Priors}.
\newblock {\em arXiv preprint arXiv:2310.12190}, 2023.

\bibitem{yan2021videogpt}
Wilson Yan, Yunzhi Zhang, Pieter Abbeel, and Aravind Srinivas.
\newblock {VideoGPT: Video Generation using VQ-VAE and Transformers}.
\newblock {\em arXiv preprint arXiv:2104.10157}, 2021.

\bibitem{yan2024forging}
Xu~Yan, Haiming Zhang, Yingjie Cai, Jingming Guo, Weichao Qiu, Bin Gao, Kaiqiang Zhou, Yue Zhao, Huan Jin, Jiantao Gao, Zhen Li, Lihui Jiang, Wei Zhang, Hongbo Zhang, Dengxin Dai, and Bingbing Liu.
\newblock {Forging Vision Foundation Models for Autonomous Driving: Challenges, Methodologies, and Opportunities}.
\newblock {\em arXiv preprint arXiv:2401.08045}, 2024.

\bibitem{yang2024generalized}
Jiazhi Yang, Shenyuan Gao, Yihang Qiu, Li~Chen, Tianyu Li, Bo~Dai, Kashyap Chitta, Penghao Wu, Jia Zeng, Ping Luo, Jun Zhang, Andreas Geiger, Yu~Qiao, and Hongyang Li.
\newblock {Generalized Predictive Model for Autonomous Driving}.
\newblock In {\em CVPR}, 2024.

\bibitem{yang2023learning}
Mengjiao Yang, Yilun Du, Kamyar Ghasemipour, Jonathan Tompson, Dale Schuurmans, and Pieter Abbeel.
\newblock {Learning Interactive Real-World Simulators}.
\newblock In {\em ICLR}, 2024.

\bibitem{yang2024video}
Sherry Yang, Jacob Walker, Jack Parker-Holder, Yilun Du, Jake Bruce, Andre Barreto, Pieter Abbeel, and Dale Schuurmans.
\newblock {Video as the New Language for Real-World Decision Making}.
\newblock {\em arXiv preprint arXiv:2402.17139}, 2024.

\bibitem{yang2024direct}
Shiyuan Yang, Liang Hou, Haibin Huang, Chongyang Ma, Pengfei Wan, Di~Zhang, Xiaodong Chen, and Jing Liao.
\newblock {Direct-a-Video: Customized Video Generation with User-Directed Camera Movement and Object Motion}.
\newblock {\em arXiv preprint arXiv:2402.03162}, 2024.

\bibitem{yang2023visual}
Zetong Yang, Li~Chen, Yanan Sun, and Hongyang Li.
\newblock {Visual Point Cloud Forecasting Enables Scalable Autonomous Driving}.
\newblock In {\em CVPR}, 2024.

\bibitem{zhai2023rethinking}
Jiang-Tian Zhai, Ze~Feng, Jinhao Du, Yongqiang Mao, Jiang-Jiang Liu, Zichang Tan, Yifu Zhang, Xiaoqing Ye, and Jingdong Wang.
\newblock {Rethinking the Open-Loop Evaluation of End-to-End Autonomous Driving in nuScenes}.
\newblock {\em arXiv preprint arXiv:2305.10430}, 2023.

\bibitem{zhang2024language}
Alex Zhang, Khanh Nguyen, Jens Tuyls, Albert Lin, and Karthik Narasimhan.
\newblock {Language-Guided World Models: A Model-based Approach to AI Control}.
\newblock {\em arXiv preprint arXiv:2402.01695}, 2024.

\bibitem{zhang2023learning}
Lunjun Zhang, Yuwen Xiong, Ze~Yang, Sergio Casas, Rui Hu, and Raquel Urtasun.
\newblock {Learning Unsupervised World Models for Autonomous Driving via Discrete Diffusion}.
\newblock In {\em ICLR}, 2024.

\bibitem{zhang2023i2vgen}
Shiwei Zhang, Jiayu Wang, Yingya Zhang, Kang Zhao, Hangjie Yuan, Zhiwu Qin, Xiang Wang, Deli Zhao, and Jingren Zhou.
\newblock {I2VGen-XL: High-Quality Image-to-Video Synthesis via Cascaded Diffusion Models}.
\newblock {\em arXiv preprint arXiv:2311.04145}, 2023.

\bibitem{zhao2024drivedreamer}
Guosheng Zhao, Xiaofeng Wang, Zheng Zhu, Xinze Chen, Guan Huang, Xiaoyi Bao, and Xingang Wang.
\newblock {DriveDreamer-2: LLM-Enhanced World Models for Diverse Driving Video Generation}.
\newblock {\em arXiv preprint arXiv:2403.06845}, 2024.

\bibitem{zhao2024unipc}
Wenliang Zhao, Lujia Bai, Yongming Rao, Jie Zhou, and Jiwen Lu.
\newblock {UniPC: A Unified Predictor-Corrector Framework for Fast Sampling of Diffusion Models}.
\newblock In {\em NeurIPS}, 2023.

\bibitem{zheng2023occworld}
Wenzhao Zheng, Weiliang Chen, Yuanhui Huang, Borui Zhang, Yueqi Duan, and Jiwen Lu.
\newblock {OccWorld: Learning a 3D Occupancy World Model for Autonomous Driving}.
\newblock {\em arXiv preprint arXiv:2311.16038}, 2023.

\bibitem{zheng2024genad}
Wenzhao Zheng, Ruiqi Song, Xianda Guo, and Long Chen.
\newblock {GenAD: Generative End-to-End Autonomous Driving}.
\newblock {\em arXiv preprint arXiv:2402.11502}, 2024.

\bibitem{zhou2024embodied}
Yunsong Zhou, Linyan Huang, Qingwen Bu, Jia Zeng, Tianyu Li, Hang Qiu, Hongzi Zhu, Minyi Guo, Yu~Qiao, and Hongyang Li.
\newblock {Embodied Understanding of Driving Scenarios}.
\newblock {\em arXiv preprint arXiv:2403.04593}, 2024.

\end{thebibliography}
}

\clearpage
\appendix
\noindent\textbf{\LARGE Appendix}
\vskip8pt
\startcontents
{
\hypersetup{linkcolor=black}
\printcontents{}{1}{}
}
\clearpage

\section{Discussions}
\label{sec:discuss}
To help a thorough understanding of this work, we discuss intuitive questions that might be raised.

\bigskip
\noindent\textbf{Q1.} \textit{Why is at least position, velocity, and acceleration required to predict coherent futures?}
\smallskip

Position ensures the predicted future begins continuously with the current state. Velocity manifests how objects are moving, \eg, whether they are turning left or turning right. Acceleration represents how velocity changes over time, \eg, whether the surroundings are moving faster or moving slower. Without utilizing acceleration as a cue, a car overtaking the ego-vehicle may suddenly be passed by in the next autoregressive prediction step. These three priors provide essential cues to allow consistent future extension with respect to historical observation.

\bigskip
\noindent\textbf{Q2.} \textit{How is the form of the proposed reward function defined?}
\smallskip

Unlike VIPER~\cite{escontrela2024video} and Diffusion Reward~\cite{huang2023diffusion} that both make discrete predictions, our model predicts continuous latent. Therefore, our reward is estimated according to conditional variance rather than log-likelihood or entropy. In addition, measuring uncertainty with log-likelihood requires comparing the prediction to the ground truth. As we deploy the reward in any scenario, the approach of VIPER is infeasible for our objective. Note that our reward calculation is meticulously designed to satisfy the Kolmogorov axioms, \ie it is non-negative and the measure of the entire sample space is $[0, 1]$.

\bigskip
\noindent\textbf{Q3.} \textit{Reward estimation efficiency compared to the detector-based method}~\cite{wang2023driving}\textit{.}
\smallskip

Though our reward estimation involves multi-round denoising, it will not spend more compute than the detector-based reward function defined in Drive-WM~\cite{wang2023driving}. To be specific, Drive-WM obtains the rewards from the perception results. Given that the detectors~\cite{li2022bevformer,liao2022maptr} take image sequences as inputs, Drive-WM has to accomplish all steps of the denoising process before perception. Differently, our reward function estimates the reward with the uncertainty that originates from the world model itself without relying on other perception models. Therefore, the estimation of uncertainty does not require completing the generation process. It can be realized by only denoising each sample for a few steps. In fact, as specified in \cref{sec:implementation}, the total computation required for reward estimation per situation (10 steps, 5 rounds) is no greater than that of generating the entire video (50 steps for our model) as Drive-WM does. Note that the computational cost for our reward estimation can be flexibly reduced to further improve its efficiency. As shown in \cref{fig:sensitivity}, using 5 denoising steps ($50\%$ of the default computation) also yields satisfactory estimations of the reward.

\bigskip
\noindent\textbf{Q4.} \textit{Usage of the proposed reward function.}
\smallskip

\textbf{(1)}~As mentioned in \cref{sec:application}, the proposed reward function can potentially serve as an alternative metric of driving actions that mitigate the concerns in existing open-loop evaluation~\cite{chen2023end,li2023ego,zhai2023rethinking}. \textbf{(2)}~As demonstrated in \cref{fig:reward}, better actions generally yield higher rewards with our reward function. Taking advantage of this property, there is great promise for our reward function to be used as a critic module~\cite{lecun2022path}, which enables model-predictive control by executing the optimal action that maximizes the estimated reward~\cite{ebert2018visual,finn2017deep,gupta2022maskvit}. This procedure can be performed in conjunction with distribution-based planners~\cite{hu2021fiery,zheng2024genad} that can make action proposals to reduce the searching space.

\bigskip
\noindent\textbf{Q5.} \textit{Any other potential applications of \modelname?}
\smallskip

\textbf{(1)}~As a generalizable predictive model, \modelname could be utilized as a forward dynamics model~\cite{cen2024using,du2023video} to simulate short-term dynamics and assist long-horizon planning tasks like visual navigation~\cite{shah2023vint}. \textbf{(2)}~It is also intriguing to utilize \modelname as an implicit driving policy, which is spontaneously acquired through future prediction~\cite{ajay2024compositional,du2024learning}. After synthesizing the video plan, we can convert the resultant image trajectory to executable actions by a non-causal inverse dynamics model~\cite{black2023zero,jordan1992forward,ko2023learning}, which can be efficiently learned from much fewer data compared with the imitation learning pipeline~\cite{baker2022video,bruce2024genie}. In autonomous driving, the inverse dynamics model could be implemented with visual odometry~\cite{lai2023xvo}. \textbf{(3)}~In collaboration with the reward function, it is also worth investigating if \modelname could facilitate model-based reinforcement learning by boosting the sampling efficiency in real-world scenarios~\cite{yang2023learning}.

\bigskip
\noindent\textbf{Q6.} \textit{Differences with GenAD}~\cite{yang2024generalized}\textit{.}
\smallskip

These two works have fundamental differences in control versatility and prediction fidelity. First of all, \modelname is a generalizable world model that can be controlled by multi-modal action conditions. Although GenAD has also trained a trajectory-conditioned extension, its weights are fully finetuned on nuScenes and the generalization of its action control has never been validated. In contrast, \modelname integrates versatile action controllability that can generalize to new scenarios in a zero-shot manner. Unlike GenAD that requires labeling OpenDV-YouTube with commands and texts, our collaborative training strategy skillfully averts this labor that may incur accumulated noises and conflicts~\cite{yang2024generalized}. In addition, \modelname (10 Hz, 576$\times$1024) operates at much higher frame rate and resolution, considerably beyond the capability of GenAD (2 Hz, 256$\times$448) in both temporal and spatial axes. Different from GenAD, we also put forth several dedicated designs for high-fidelity prediction. We find that \modelname, with a lower model complexity, achieves much better FID and FVD scores than GenAD (see \Cref{tab:fvd}).

\bigskip
\noindent\textbf{Q7.} \textit{Limitations, failure cases, and possible solutions.}
\smallskip

As one of the pioneering efforts, \modelname still has a few limitations that call for future works. \textbf{(1)}~Since \modelname predicts futures at an exceptional spatiotemporal resolution, it is inevitable to be computationally expensive, particularly in downstream applications. Potential solutions may include faster sampling techniques~\cite{lu2022dpm,zhao2024unipc} and training-based distillations~\cite{luo2023latent,meng2023distillation,salimans2022progressive,sauer2024fast,song2023consistency,wang2023videolcm}. \textbf{(2)}~It is possible that the prediction may undergo an apparent degradation in long-horizon rollouts or drastic view shifts. Extra refinements on the prediction results~\cite{blattmann2023align,ho2022cascaded,hu2023gaia,podell2023sdxl,wang2023lavie,zhang2023i2vgen} could be helpful. Speculatively, applying our recipe to more scalable architecture~\cite{hu2023gaia,peebles2023scalable} is also promising to address this limitation. \textbf{(3)}~Similar to other controllable video generation methods~\cite{wang2023motionctrl}, the chance of failure still persists in our action controls, especially for ambiguous intentions such as commands and goal points as \cref{fig:action_effect} reveals. Incorporating more datasets with action annotations~\cite{caesar2021nuplan,sun2020scalability,wilson2023argoverse} for collaborative training could be beneficial. Using compositional classifier-free guidance~\cite{brooks2023instructpix2pix,chen2023gentron,girdhar2023emu} to amplify the individual impact of action conditions may also help (at a cost of increased inference compute). \textbf{(4)}~Although our training data is based on the largest public driving dataset~\cite{yang2024generalized}, it is nowhere near the entirety of Internet driving data, thus leaving huge untapped potential to further expand the capabilities of \modelname.

\bigskip
\noindent\textbf{Q8.} \textit{Why not expand the \modelname framework to surround-view generation?}
\smallskip

It is true that supporting surround-view generation would further help driving. Existing works~\cite{wang2023driving,zhao2024drivedreamer} have explored the surround-view settings on nuScenes~\cite{caesar2020nuscenes}. However, in this paper, we focus on the front-view setting for three main reasons: \textbf{(1)}~The front view setting allows leveraging diverse data sources~\cite{hu2023gaia,yang2024generalized}. Conversely, the distinctions in multi-view videos from various datasets, such as different numbers of cameras, hinder unified modeling and data scaling. \textbf{(2)}~Models that focus on the front view can be seamlessly applied to different datasets without adaptation~\cite{sima2023drivelm}, broadening their applicability across datasets. \textbf{(3)}~Though incomplete, the front view often contains most of the information vital for driving. As demonstrated in NAVSIM~\cite{dauner2024navsim}, using the front view alone results in only a $1.1\%$ performance downtick in collision rate compared to using five surround-view cameras.

\bigskip
\noindent\textbf{Q9.} \textit{Broader impact.}
\smallskip

Despite the encouraging improvements, our work is by no means perfect when it comes to real-world applications that involve dealing with highly complicated situations. As \modelname is based on the diffusion framework, which introduces stochastic outcomes and non-negligible latencies, deploying it into autonomous vehicles directly could pose safety risks. While it is not a silver bullet yet, we expect that \modelname will inspire the community to further exploit the capabilities and applications of driving world models. As a prototype for generalizable driving world models, we hope that \modelname can stimulate the investigations in developing generalizable systems for autonomous driving and machine intelligence.

\section{Related Work}
\label{sec:related}

\subsection{World Models}
\vspace{-0.5mm}
Intelligent agents should be able to make efficacious decisions even under unseen situations~\cite{bruce2024genie,hu2023toward,lecun2022path,sutton2022quest,yang2024video,zhou2024embodied}. This requires fundamental knowledge of the world that generalizes to rare cases. As an internal manifestation of such knowledge, a world model predicts plausible futures of the world given potential actions~\cite{bruce2024genie,hafner2019dream,kim2020learning,lecun2022path,oh2015action,sutton2022quest,yang2023learning}. In principle, it not only predicts how the environment will unfold over time, but also deduces the underlying physical dynamics and agentic behaviors. Such properties can be useful for representation learning~\cite{gupta2022maskvit,he2024large,lotter2016deep,schwarzer2020data,wu2024pre}, model-based reinforcement learning~\cite{ha2018recurrent,hafner2019dream,hafner2020mastering,hafner2023mastering,micheli2022transformers,oh2015action,piergiovanni2019learning}, as well as model-predictive control~\cite{ebert2018visual,finn2017deep,gupta2022maskvit,hafner2019learning,mendonca2023structured,schubert2023generalist,zhang2024language}. Recent methods~\cite{gurnee2023language,hao2023reasoning,li2022emergent,lin2023learning,liu2024world} also induce language-based world models from large language models, but are restricted in textual space and struggle with grounding on physics~\cite{du2023video,hu2023language}.

Although world models have been extensively applied and made significant revolutions in simulated games~\cite{hafner2019dream,hafner2020mastering,hafner2023mastering} and indoor embodiment~\cite{koh2021pathdreamer,mendonca2023structured,wang2023dreamwalker}, such investigations for autonomous driving still lag behind~\cite{wang2023driving,zhang2023learning}. Different from other tasks, world modeling for autonomous driving poses unique challenges, which primarily arise from the large field of views with highly dynamic motions. Some practices imagine the world in the bird's eye view (BEV) space~\cite{chitta2024sledge,gao2022enhance,hu2022model,hu2021fiery,li2023delving,li2024think2drive}. Recent practices model the world state as raw sensor observations such as point clouds~\cite{bogdoll2023muvo,khurana2023point,yang2023visual,zhang2023learning,zheng2023occworld} and images~\cite{hu2023gaia,jia2023adriver,kim2021drivegan,lu2023wovogen,santana2016learning,wang2023drivedreamer,wang2023driving,wu2024pre,zhao2024drivedreamer}. The latter category, namely visual world models, hold more promise for scaling up due to sensor flexibility and data accessibility. Nevertheless, existing methods are restricted to a particular dataset~\cite{jia2023adriver,li2023open,lu2023wovogen,wang2023drivedreamer,wang2023driving,zhang2023learning,zhao2024drivedreamer,zheng2023occworld} or simulator~\cite{bogdoll2023muvo,wu2024pre}, compromising their generalization ability to novel domains. Meanwhile, these efforts lack systematic designs for the driving domain and only model the world at relatively low frame rates and resolutions, which discards the fine-grained details and impairs their ability to express real-world behaviors. Moreover, most of them are restricted to a specific control modality~\cite{hu2023gaia,jia2023adriver,lu2023wovogen,wang2023drivedreamer}, which hinders the accommodation to prevailing planning algorithms~\cite{casas2021mp3,chen2020learning,chitta2022transfuser,hu2022st,hu2023planning,jiang2023vad} and extension to more applications like decision-making~\cite{wang2023driving} or user interaction~\cite{kong2023dreamdrone}. Besides, existing methods seldom explore zero-shot action controllability across different datasets. The inferior generalization, fidelity and controllability collectively preclude existing driving world models from broadly facilitating the development of autonomous driving.

\subsection{Video Generation}
\vspace{-0.5mm}
Video generation is an effective way to model the world and has undergone remarkable advancements over the years. Pioneering works~\cite{vondrick2016generating,yan2021videogpt} have studied various kinds of generative models~\cite{esser2021taming,goodfellow2014generative,kingma2013auto}. Swayed by the success of diffusion models~\cite{dhariwal2021diffusion,ho2020denoising,rombach2022high}, a surge of diffusion-based video generation methods have emerged~\cite{blattmann2023align,guo2023animatediff,he2022latent,ho2022video,singer2022make,voleti2022mcvd,wang2023modelscope}. Recent works~\cite{blattmann2023stable,chen2023videocrafter1,girdhar2023emu,zhang2023i2vgen} shift their focus towards image-to-video generation for its finer content description and better scalability in training data. However, most of them are not strict predictive models that generate videos starting from the condition image. Moreover, existing methods struggle with the intricate dynamics in driving scenarios from the ego perspective~\cite{yang2024generalized}, which limits their feasibility as driving world models.

While the majority of existing methods produce videos without explicit controllability, two recent works~\cite{wang2023motionctrl,yang2024direct} introduce camera motion control to video generation. However, camera motion is conceptually distinct from vehicle actions and both of these works are text-to-video methods without any prediction ability. Contrarily, the model we developed is a predictive world model that produces realistic dynamics and allows versatile action controls for autonomous driving.

\section{Implementation Details}
\label{sec:implementation}

\subsection{Model}
\vspace{-0.5mm}
We adopt the framework of SVD~\cite{blattmann2023stable} as the architecture of \modelname, which consists of 2.5B parameters in total, including 1.6B UNet parameters. For action conditioning, we encode the value of each action sequence into Fourier embeddings~\cite{tancik2020fourier,vaswani2017attention} with 128 channels.

\subsection{Dataset}
\vspace{-0.5mm}
We utilize a rigorously filtered set of OpenDV-YouTube~\cite{yang2024generalized} for training, and incorporate nuScenes training set~\cite{caesar2020nuscenes} during the action control learning phase. Concretely, we manually eliminate 15 hours of irrelevant content from OpenDV-YouTube, yielding approximately 1735 hours of unlabeled driving videos. Since nuScenes is heavily biased~\cite{li2023ego,sima2023drivelm,wang2023driving}, we balance its samples based on command categories to foster the learning of rare actions. The video clips are sampled with 25 frames at 10 Hz. Although nuScenes~\cite{caesar2020nuscenes} is logged at 12 Hz, we find no negative impact of treating them as 10 Hz videos. The model inputs are composed by cropping and resizing these clips to the target resolution. We define an action as a sequence comprising 25 frames. To categorize actions into commands, we follow the established conventions in planning~\cite{hu2022st,hu2023planning,jiang2023vad} and define the command of ego-vehicle as "\texttt{turn right}" or "\texttt{turn left}" when its final displacement exceeds 2 meters in the orthogonal direction relative to its initial heading. To allow more precise categorization, we additionally introduce a "\texttt{stop}" command when the forward driving distance is less than 2 meters.

\subsection{Training}
\vspace{-0.5mm}
At the first training phase, we train all UNet parameters at 576$\times$1024 resolution on 128 A100 GPUs for 20K iterations, which takes about 8 days in total. We accumulate the gradients of 2 steps, yielding an effective batch size of 256. Following SVD, our model is trained with the EDM framework~\cite{karras2022elucidating}. We use the AdamW optimizer~\cite{loshchilov2017decoupled} with a learning rate of $1\times10^{-5}$. The learning rate for spatial layers is moderated by a discount factor of $0.1$. The coefficients $\lambda_{1}$ and $\lambda_{2}$ in \cref{eq:final} are set to $1.0$ and $0.1$ respectively. Offset noise~\cite{guttenberg2023offset} is also used with a strength of $0.02$ as it helps improve temporal smoothness. We randomly sample different orders of dynamic priors with increasing probabilities, \ie $\sfrac{1}{15}$, $\sfrac{2}{15}$, $\sfrac{4}{15}$, $\sfrac{8}{15}$ for 0, 1, 2, 3 condition frames respectively. The noise augmentation~\cite{ho2022cascaded} is disabled to retain more details from the condition frames. 

As for the action control learning phase, we freeze the pretrained weights and add LoRA~\cite{hu2021lora} and projection layers to all attention blocks of the UNet. The rank of LoRA is set to 16. We then train the new weights at 320$\times$576 resolution for 120K iterations using batch size 8 and learning rate $5\times10^{-5}$. After the controllability can be clearly witnessed, we continue to finetune the unfrozen weights at 576$\times$1024 resolution for another 10K iterations. We drop out each activated action mode with a ratio of $15\%$ to allow classifier-free guidance~\cite{ho2022classifier}. The sampling ratio of OpenDV-YouTube and nuScenes is $1:1$ at this training phase. The whole training process for action controllability takes around 10 days on 8 A100 GPUs, with roughly 8 days at the low resolution and 2 days at the high resolution.

\subsection{Sampling}
\vspace{-0.5mm}
We generate future videos using the DDIM sampler~\cite{song2020denoising} for 50 steps. The sampling starts with $\sigma_{\text{max}}$ at $700.0$. Since our model predicts long-term futures in an autoregressive manner, the issue of over-saturation caused by the standard classifier-free guidance will accumulate rapidly. Therefore, unlike SVD that linearly increases the guidance scale, we employ a triangular classifier-free guidance scheme~\cite{voleti2024sv3d} to permit genuine long-horizon rollouts. Specifically, for the $i$-th frame in each $K$ frames to predict, we assign its guidance scale $s(i)$ following:
\begin{equation}
\label{eq:triangular}
s(i)=\begin{cases}s_{\text{min}}+\frac{2i}{K}(s_{\text{max}}-s_{\text{min}})&\text{if }i<\frac{K}{2}\\s_{\text{max}}-\frac{2(K-i)}{K}(s_{\text{max}}-s_{\text{min}})&\text{if }i\geq\frac{K}{2}\end{cases},
\end{equation}
where $s_{\text{min}}$ and $s_{\text{max}}$ indicate the minimum and maximum guidance scales along the temporal axis. In our experiments, we define $s_{\text{min}}$ as $1.0$ and $s_{\text{max}}$ as $2.5$. This triangle scheme assigns moderate guidance scales to the frames that will be used as conditions in the next prediction round. Due to sufficient temporal interaction, the quality of intermediate frames can also propagate to the frames that have lower guidance scales. As illustrated in \cref{fig:scale}, this technique adeptly mitigates the saturation drift problem while enhancing details. To improve perceptual continuity, we split the generated latent into clips with an overlap of 3 frames before sending them to the video-aware decoder~\cite{blattmann2023stable}. After decoding, the overlapped frames are averaged pixel-wise.

\subsection{Human Evaluation}
\vspace{-0.5mm}
Recall that we ask the participants to judge side-by-side video pairs from visual quality and motion rationality. To guarantee credible responses, we provide detailed commentary for each aspect of the human evaluation. For visual quality, we let the participants focus on the consistency and harmony of the generated content. For motion rationality, we encourage the participants to pay more attention to the plausibility of the ways that ego-vehicle and other agents move, \eg, whether they are following the traffic rules and exhibiting safe behaviors. For all public models we compared, we use the official checkpoints and configurations for inference without finetuning. For the models that require textual inputs~\cite{xing2023dynamicrafter,zhang2023i2vgen}, we set the prompt as "\texttt{realistic drive view}".

\subsection{Reward Estimation}
\vspace{-0.5mm}
For each condition frame and action pair, we accumulate an ensemble with size $M=5$ to obtain a reliable uncertainty estimation. Each sample in the ensemble is inferred for 10 denoising steps as we find it is unnecessary to generate high-quality results for uncertainty estimation. The coefficient $\beta$ in the correlating strategy~\cite{gupta2022maskvit,nagabandi2020deep} is set to 0.5.

\subsection{Ablation Studies}
\vspace{-0.5mm}
For the ablation of loss functions, we train each variant on OpenDV-YouTube~\cite{yang2024generalized} for 10K steps at a spatial resolution of 576$\times$1024. All ablations, including the additional ablations in \cref{sec:add_exp}, are initialized by loading the pretrained checkpoint of SVD~\cite{blattmann2023stable} and conducted with 8 A100 GPUs.

\section{Additional Experiments}
\label{sec:add_exp}

\begin{table}[t!]
\vspace{-1.5mm}
\begin{minipage}[c]{0.49\linewidth}
\centering
\small\setlength{\tabcolsep}{4pt}
\caption{\textbf{Reward of commands on Waymo.} The ground truth commands generally obtain higher rewards than random command inputs, suggesting that the proposed reward function can be used as a reliable indicator for command selection.}
\label{tab:reward}
\vspace{-0.5mm}
\begin{tabular}{l|c}
\toprule
\textbf{Condition} & \textbf{Average Reward} \\
\midrule
GT Com. & 0.892 \\
random Com. & 0.878 (-0.014) \\
\bottomrule
\end{tabular}
\end{minipage}\hfill
\begin{minipage}[c]{0.49\linewidth}
\centering
\small\setlength{\tabcolsep}{4pt}
\caption{\textbf{Effect of action independence.} Without losing generality, we choose the trajectory as a representative action for evaluation. The proposed constraint expedites the learning of actions.}
\label{tab:training_strategy}
\vspace{-1mm}
\begin{tabular}{c|l|cccc}
\toprule
\multirow{2}{*}{\textbf{Strategy}} & \multicolumn{1}{c|}{\multirow{2}{*}{\textbf{Action}}} & \multicolumn{4}{c}{\textbf{Subset FVD$\downarrow$}} \\
& & forth & right & left & stop \\
\midrule
\multirow{2}{*}{\textit{w/o} A.I.} & \textit{w/o} Traj. & 163.0 & 273.9 & 428.3 & 497.1 \\
& \textit{w/} Traj. & 138.8 & 232.9 & 368.2 & 132.3 \\
\midrule
\multirow{2}{*}{\textit{w/} A.I.} & \textit{w/o} Traj. & 156.2 & 263.7 & 402.9 & 463.7 \\
& \textit{w/} Traj. & 130.7 & 230.8 & 345.7 & 118.9 \\
\bottomrule
\end{tabular}
\end{minipage}
\vspace{-5mm}
\end{table}

\begin{wrapfigure}{r}{0.33\textwidth}
\centering
\vspace{-15mm}
\includegraphics[width=0.33\textwidth]{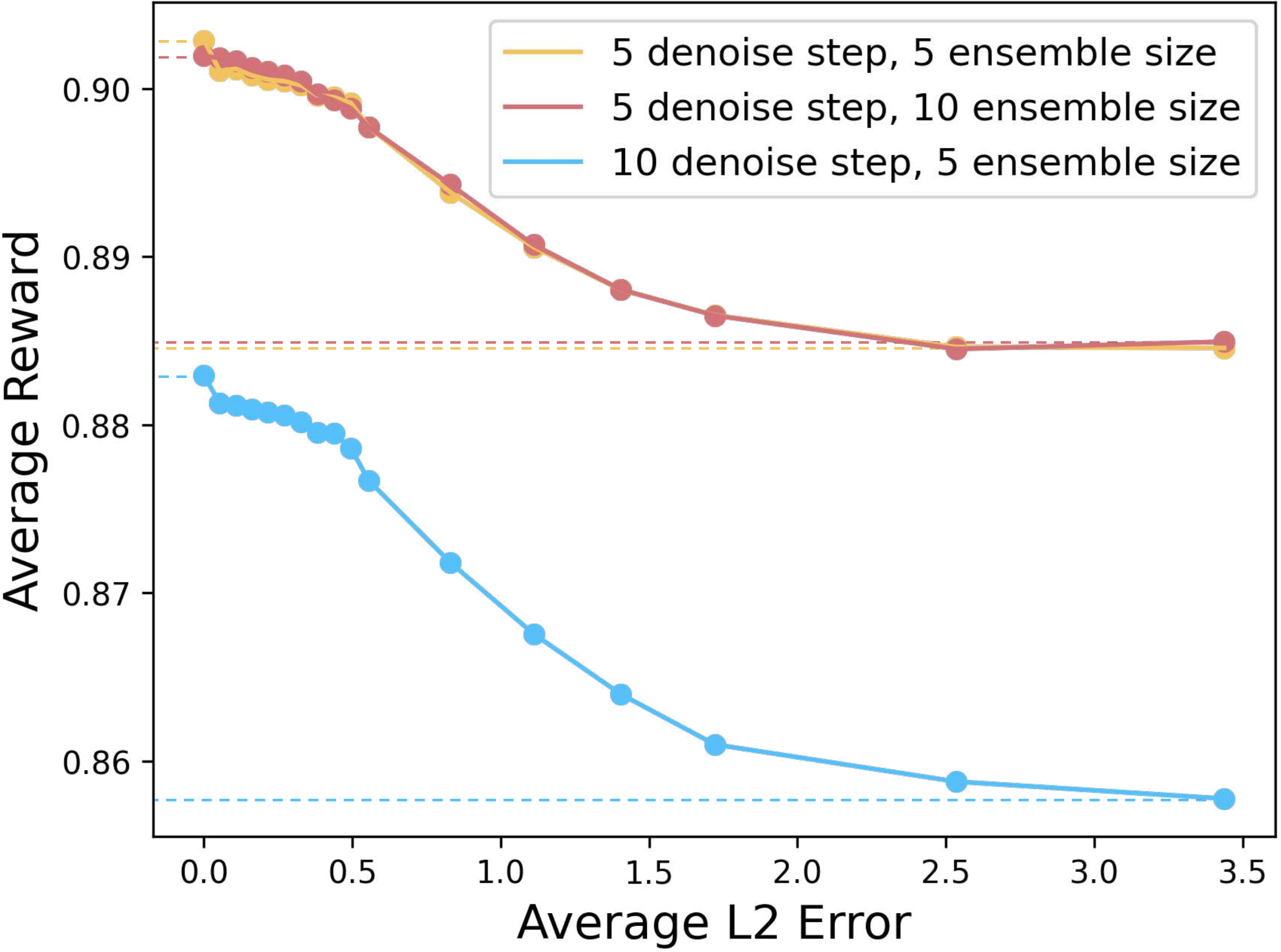}
\vspace{-5mm}
\caption{\textbf{Sensitivity of reward estimation to hyperparameters.} Increasing the number of denoising steps can produce more discriminative rewards, whereas increasing the ensemble size can slightly stabilize the estimations.}
\label{fig:sensitivity}
\vspace{-10mm}
\end{wrapfigure}

\subsection{Parameter Sensitivity of Reward Estimation}
\vspace{-0.5mm}
To investigate how the number of denoising steps and the ensemble size influence the performance of the proposed reward function, we repeat the reward estimation procedure in \cref{sec:application} with different hyperparameter settings. We start off by using 5 denoising steps and an ensemble size of 5 for each sample. We then test two variants that double the computational cost by increasing the denoising steps to 10 (our default setup in \cref{sec:implementation}) and increasing the ensemble size to 10 respectively. Following \cref{sec:application}, we plot the correlation of the estimated rewards with L2 errors for the three variants in \cref{fig:sensitivity}. The results show that increasing the number of denoising steps can greatly enlarge the relative contrast of rewards, suggesting that denoising step is a more important factor than ensemble size under the same computation budget for reward estimation.

\subsection{Reward Estimation for Commands}
\vspace{-0.5mm}
To demonstrate that the proposed reward function is also applicable for other actions, we estimate the rewards of ground truth commands from Waymo and compare them with the rewards of random commands. The results in \Cref{tab:reward} suggest that our reward is also competent for command selection.

\subsection{Action Independence Constraint}
\vspace{-0.5mm}
To prove the efficacy of our learning strategy for action control, we conduct a comparison by removing the action independence constraint proposed in \cref{sec:action}. We train two variants on nuScenes~\cite{caesar2020nuscenes} at the resolution of 320$\times$576 pixels for 62K steps. The comparison results are presented in \Cref{tab:training_strategy}.

\begin{figure}[t!]
\centering
\includegraphics[width=0.99\textwidth]{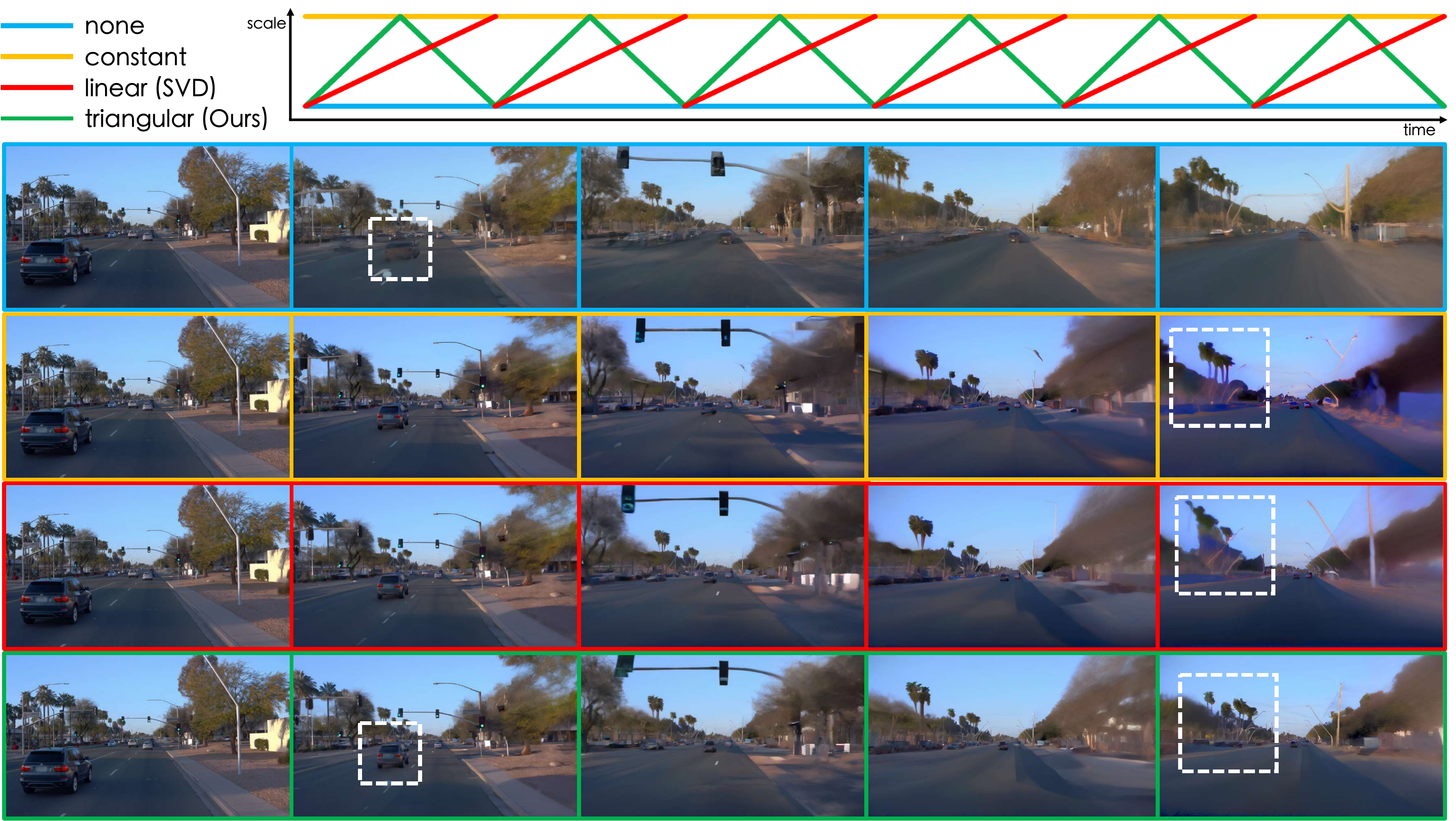}
\vspace{-2mm}
\caption{\textbf{Effect of guidance scale.} We predict 15s long-term videos with different CFG schemes. Our method achieves the optimal equilibrium between detail generation and saturation maintenance.}
\label{fig:scale}
\vspace{-2mm}
\end{figure}

\begin{figure}[t!]
\centering
\includegraphics[width=0.99\textwidth]{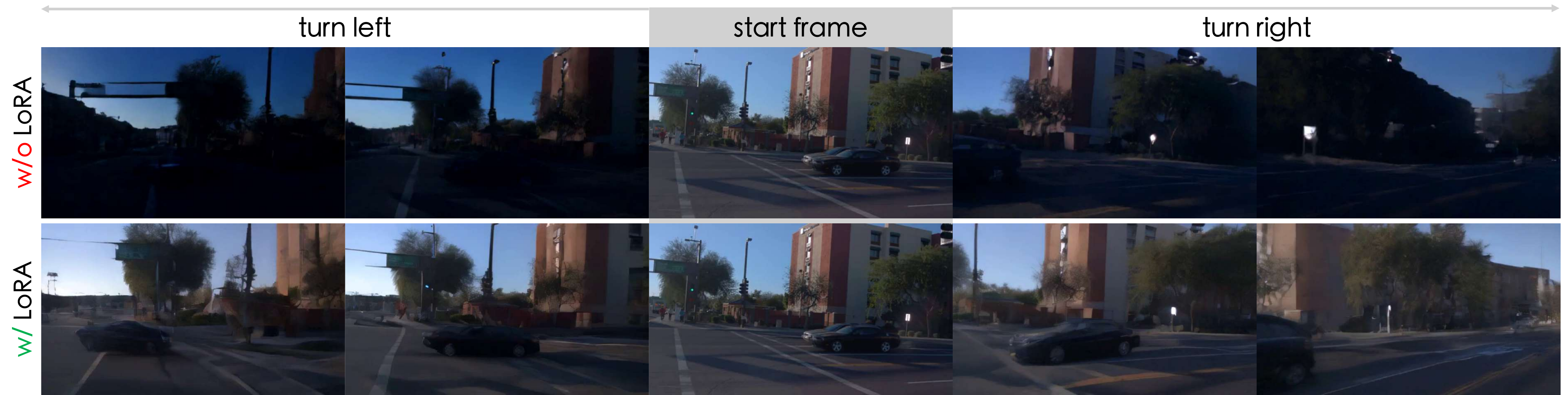}
\vspace{-2mm}
\caption{\textbf{Necessity of LoRA adaptation.} Training newly added projections alone without LoRA results in visual corruptions. The compared variants are trained on nuScenes and inferred on Waymo.}
\label{fig:lora}
\vspace{-2mm}
\end{figure}

\subsection{Triangular Guidance Scheme}
\vspace{-0.5mm}
We further compare the introduced classifier-free guidance scheme with the vanilla scheme and the linear scheme~\cite{blattmann2023stable} to verify its necessity. \cref{fig:scale} shows that our triangular scaling attains the best trade-off between visual quality and saturation preservation.

\subsection{LoRA Adaptation}
\vspace{-0.5mm}
To show the necessity of applying LoRA in \cref{sec:action}, we train two variants at a resolution of 320$\times$576 pixels for 30K iterations. With the pretrained UNet weights fixed, we let one variant train LoRA and action projection layers in the attention blocks, while the other adjusts new projection layers only. As shown in \cref{fig:lora}, adding LoRA is essential for action control learning.

\subsection{Action Control Consistency}
\vspace{-0.5mm}
In \Cref{tab:rebuttal_fvd}, we report the complete FVD scores of \cref{fig:action_effect}, which further validates the effectiveness of all kinds of action controls. Note that since our "\texttt{stop}" subset consists of samples where the final displacements are within 2 meters, the goal points typically do not appear in these videos. Hence, for the experiment that uses goal point as action condition on the "\texttt{stop}" subset, most samples are generated in the same way as the action-free mode.

\subsection{Human Evaluation with GenAD}
\vspace{-0.5mm}
To our best knowledge, there is not a driving-specific world model publicly available so far, making it hard to conduct qualitative human evaluation. Hence, we mainly compare \modelname against the existing methods with the officially reported FID and FVD scores in \Cref{tab:fvd}.

\begin{wrapfigure}{r}{0.33\textwidth}
\centering
\vspace{-6mm}
\includegraphics[width=0.28\textwidth]{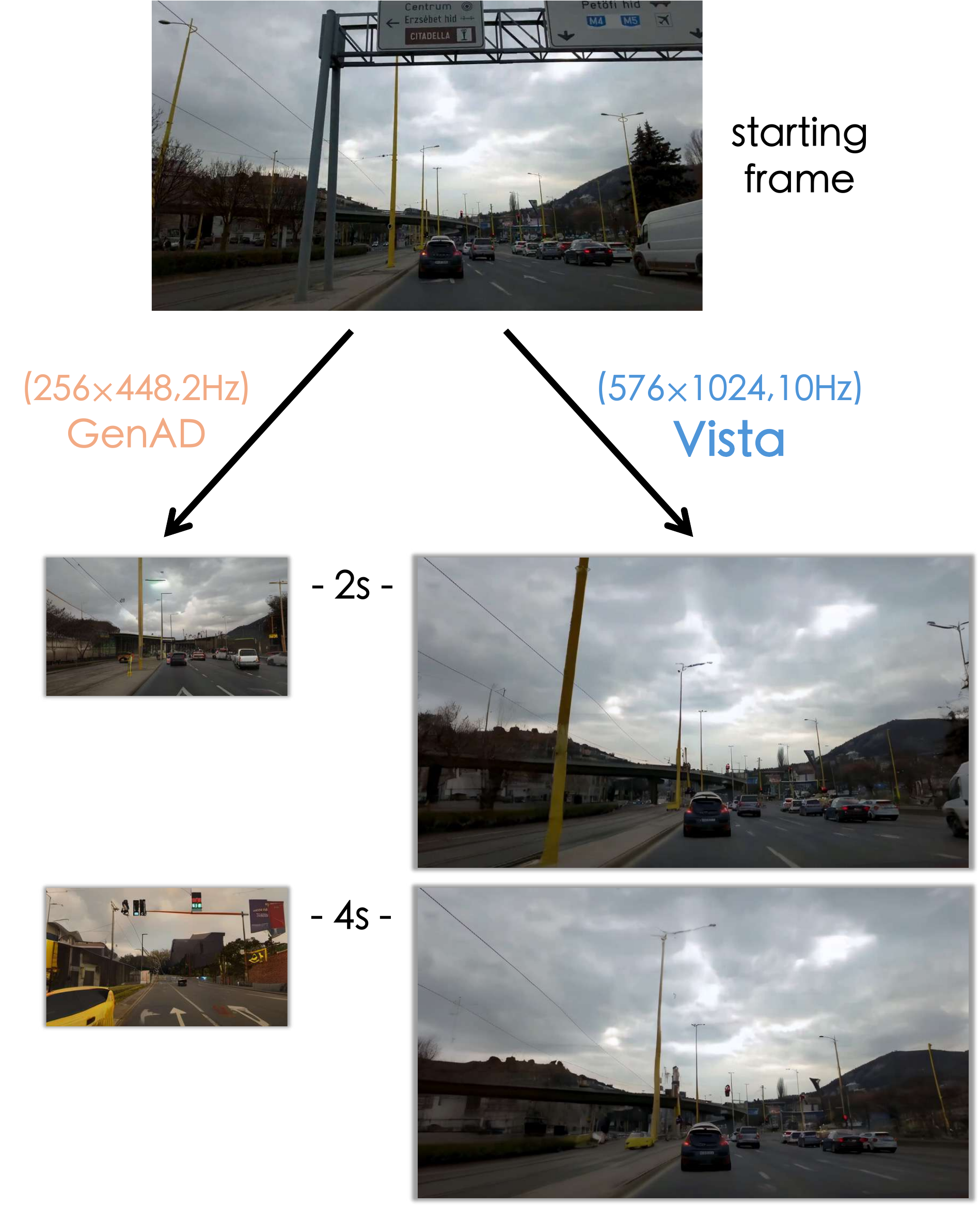}
\vspace{-1mm}
\caption{\textbf{Perceptual disparity between GenAD and \modelname.}}
\label{fig:res}
\vspace{-3mm}
\end{wrapfigure}

To demonstrate the considerable improvements in visual quality and motion rationality, we conduct an extra human evaluation with the state-of-the-art GenAD model~\cite{yang2024generalized}. Since GenAD processes a 4-second video each time, we perform autoregressive prediction to extend \modelname’s output to 5 seconds and trim the last second to align with GenAD’s duration. To avoid any bias caused by resolution and frequency, we downsample the outputs of \modelname (576$\times$1024 resolution at 10 Hz) to 256$\times$448 resolution at 2 Hz. The evaluation follows the same procedure specified in \cref{sec:evaluation}.

We collect 25 diverse samples from the unseen OpenDV-YouTube-val set and invite 20 volunteers for evaluation. We ask the volunteers to choose the video they deem better. As a result, \modelname is preferred in $94.4\%$ and $94.8\%$ of the time on visual quality and motion rationality respectively. This attests that \modelname, in spite of experiencing a large perceptual reduction due to downsampling, still exhibits a significant advantage over GenAD in generation quality. We also compare the predictions of GenAD and \modelname in \cref{fig:res}, showing the superiority of \modelname in resolution and fidelity.

\begin{table}[t!]
\caption{\textbf{Complete FVD scores of different action categories.} We obtain the FVD scores on four subsets divided by command categories. All types of action controls are effective across all categories.}
\vspace{-1mm}
\label{tab:rebuttal_fvd}
\centering
\scalebox{0.8}{
\begin{tabular}{>{\centering}p{0.14\textwidth} | p{0.20\textwidth} | >{\centering}p{0.14\textwidth} >{\centering}p{0.14\textwidth} >{\centering\arraybackslash}p{0.14\textwidth} >{\centering\arraybackslash}p{0.14\textwidth} >{\centering\arraybackslash}p{0.14\textwidth}}
\toprule
\multirow{2}{*}{\textbf{Dataset}} & \multicolumn{1}{c|}{\multirow{2}{*}{\textbf{Condition}}} & \multicolumn{5}{c}{\textbf{Subset FVD $\downarrow$}} \\
& & forth & right & left & stop & average \\
\midrule
\multirow{5}{*}{nuScenes} & action-free & 135.6 & 405.6 & 513.8 & 414.1 & 367.2 \\
& + goal point & 122.4 & 315.6 & 439.6 & 413.5 & 322.7 \\
& + command & 122.2 & 299.7 & 485.6 & 261.6 & 292.2 \\
& + angle \& speed & 122.8 & 285.6 & 397.8 & 114.1 & 230.0 \\
& + trajectory & 125.2 & 229.2 & 357.7 & 118.5 & 207.6 \\
\midrule
\multirow{3}{*}{Waymo} & action-free & 145.9 & 407.6 & 529.9 & 164.1 & 311.8 \\
& + command & 122.5 & 331.5 & 496.9 & 143.9 & 273.7 \\
& + trajectory & 126.3 & 285.5 & 527.6 & 136.5 & 268.9 \\
\bottomrule
\end{tabular}
}
\vspace{-2mm}
\end{table}

\section{Additional Visualizations}
\label{sec:add_vis}

\subsection{Generalization Ability}
\begin{figure}[t!]
\centering
\includegraphics[width=0.99\textwidth]{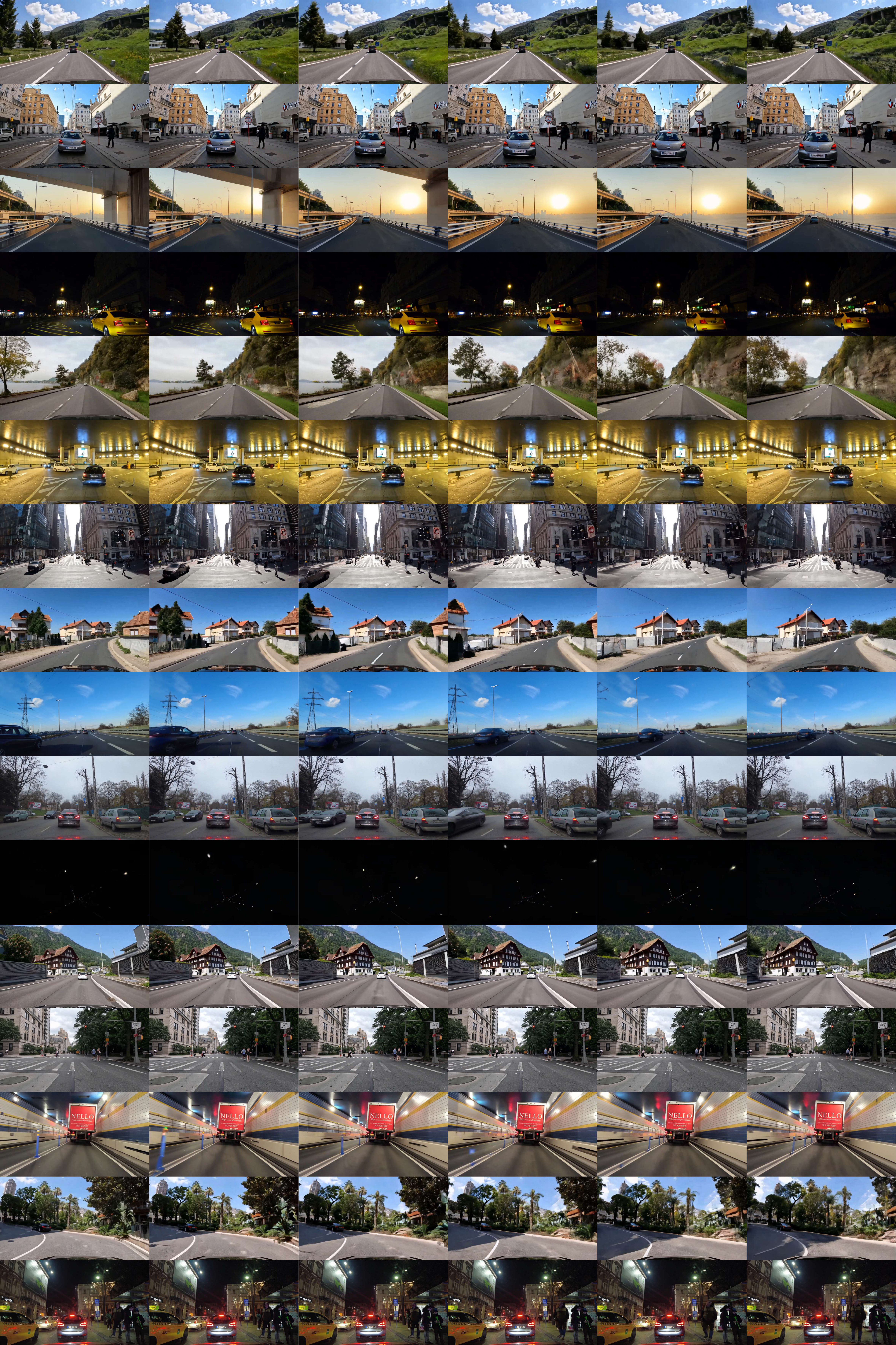}
\vspace{-2mm}
\caption{\textbf{Generalization ability of \modelname.} We apply \modelname across diverse scenes (\eg, countrysides and tunnels) with unseen camera poses (\eg, the perspective of a double-decker bus). Our model can predict high-resolution futures with vivid behaviors of vehicles and pedestrians, exhibiting strong generalization abilities and profound comprehension of world knowledge. Best viewed zoomed in.}
\label{fig:open_world_short_term}
\vspace{-2mm}
\end{figure}

\begin{figure}[t!]
\centering
\includegraphics[width=0.99\textwidth]{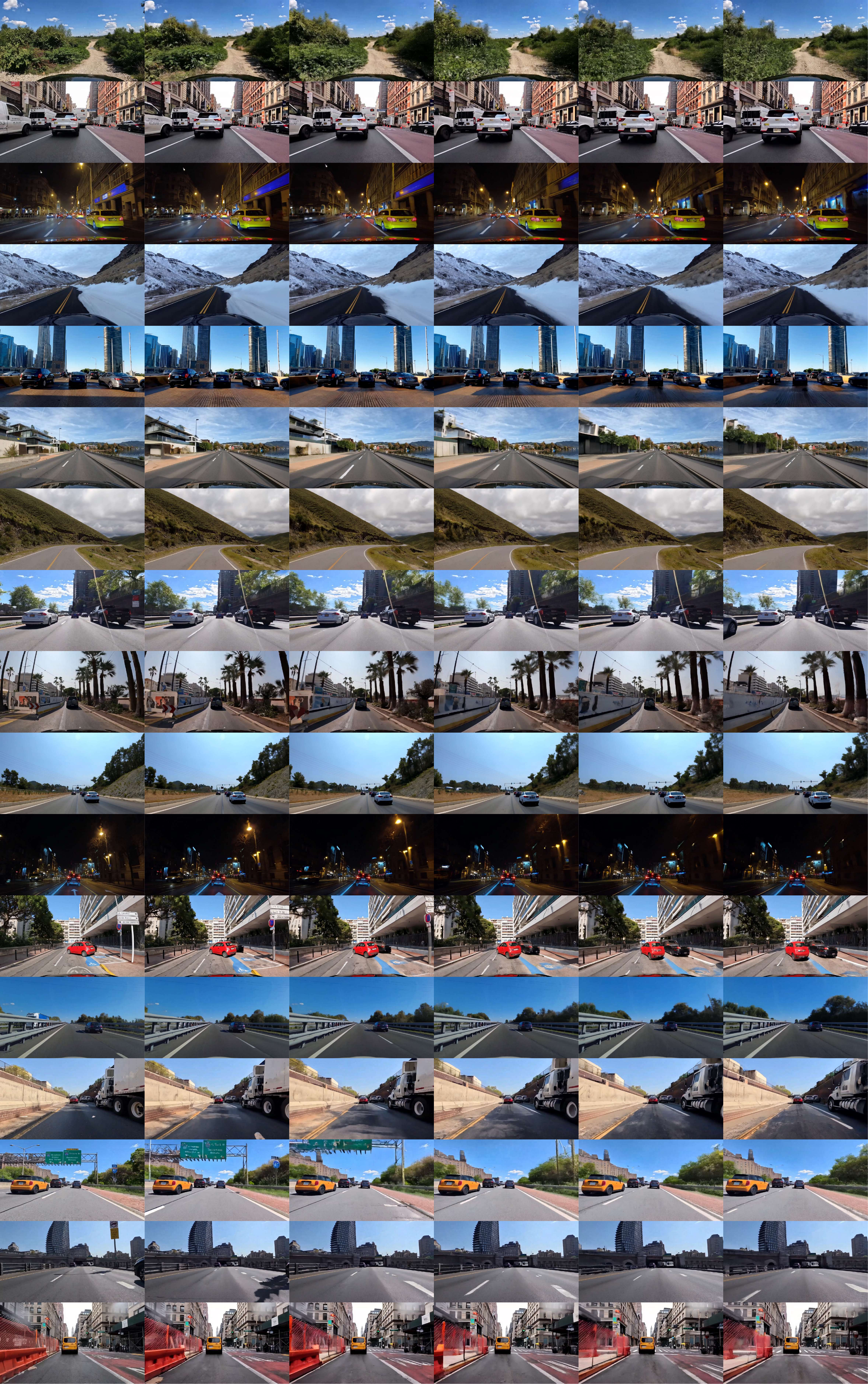}
\vspace{-2mm}
\caption{\textbf{Generalization ability of \modelname in more scenarios.} Best viewed zoomed in.}
\label{fig:open_world_short_term2}
\vspace{-2mm}
\end{figure}

\vspace{-0.5mm}
We further demonstrate the strong generalization ability of \modelname by deploying it to different scenarios in the wild. The results in \cref{fig:open_world_short_term} and \cref{fig:open_world_short_term2} illustrate that \modelname can make high-fidelity predictions in a very diverse range of scenarios.

\subsection{Long-Horizon Prediction}
\begin{figure}[t!]
\centering
\includegraphics[width=0.99\textwidth]{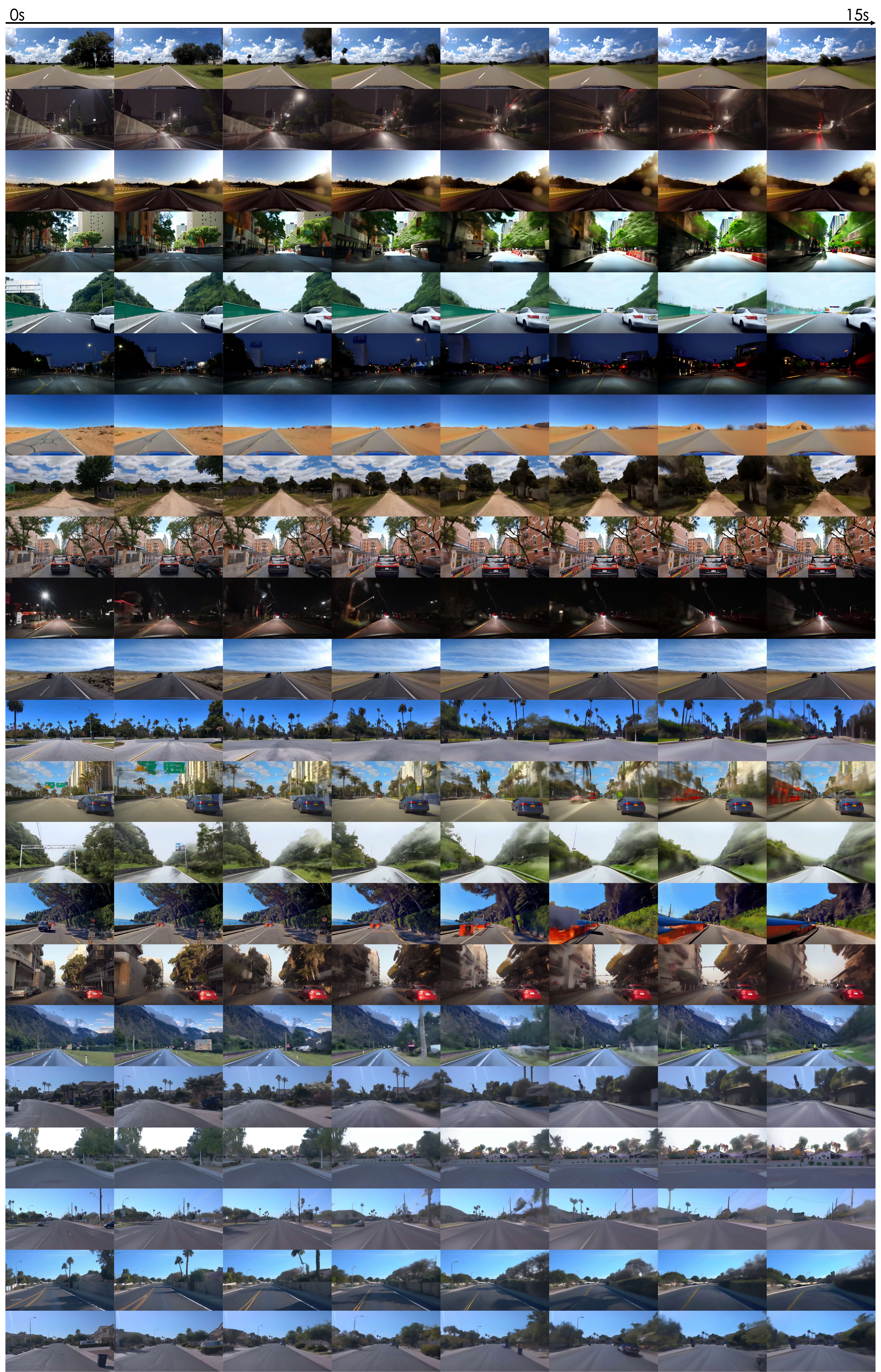}
\vspace{-2mm}
\caption{\textbf{Additional results of long-horizon prediction.} Our model can autoregressively simulate long driving experiences with marginal quality decline. All videos continue for 15 seconds at 10 Hz.}
\label{fig:long_more}
\vspace{-2mm}
\end{figure}

\vspace{-0.5mm}
In addition to \cref{fig:long}, we provide more qualitative visualizations of long-horizon prediction in \cref{fig:long_more}. \modelname can continuously predict long-term futures with consistent content and motion.

\subsection{Action Controllability}
\begin{figure}[t!]
\centering
\includegraphics[width=0.99\textwidth]{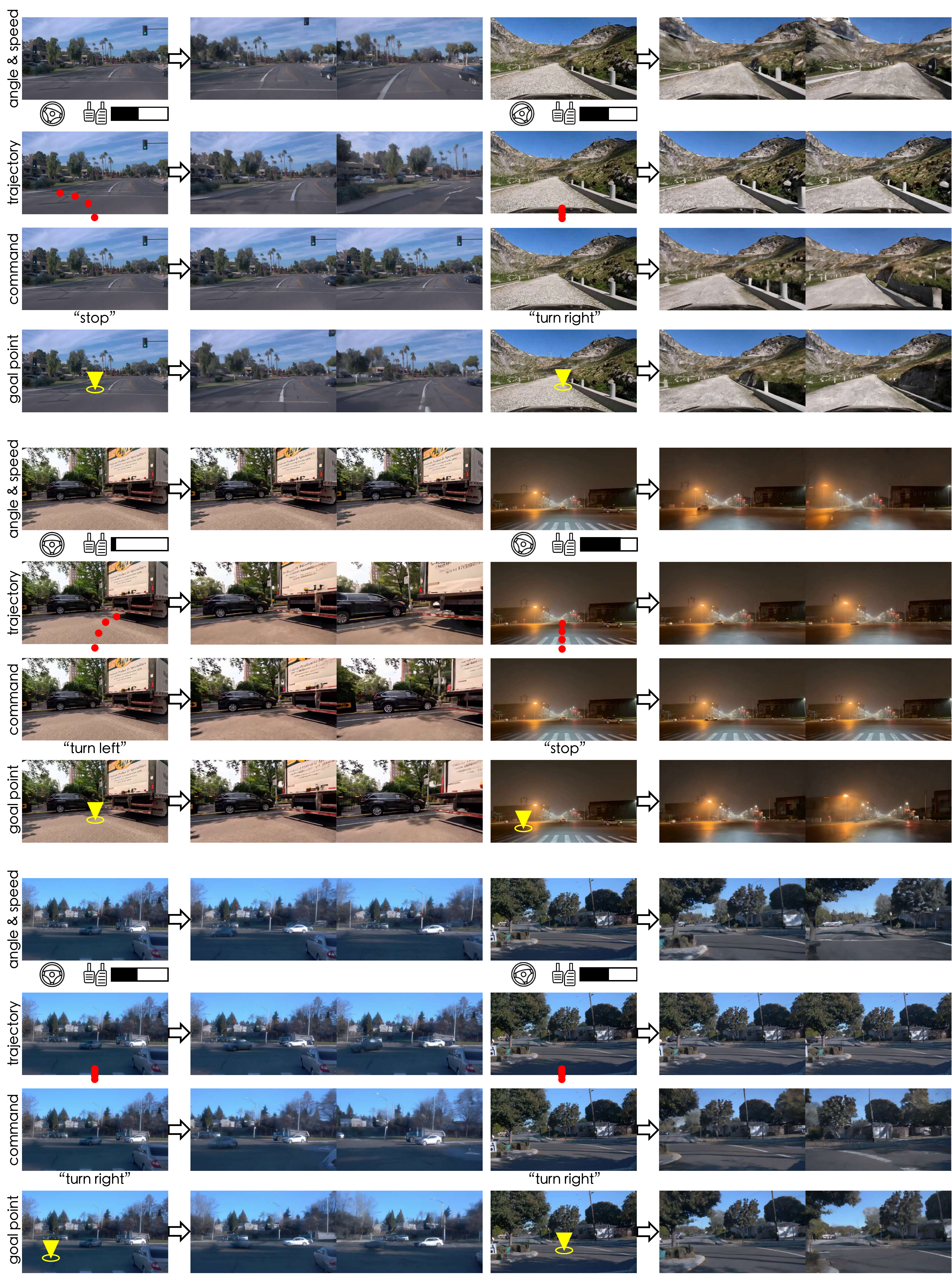}
\vspace{-2mm}
\caption{\textbf{Additional results of action controllability.} We trial different action conditions across multiple scenes from OpenDV-YouTube-val and Waymo. The behaviors of the ego-vehicle can be consistently controlled by various kinds of interventions.}
\label{fig:action_more}
\vspace{-2mm}
\end{figure}

\vspace{-0.5mm}
We provide more prediction results with different action inputs in \cref{fig:action_more}. The results on OpenDV-YouTube-val~\cite{yang2024generalized} and Waymo~\cite{sun2020scalability} show that the versatile controllability of \modelname can be readily transferred to different domains in a zero-shot manner.

\subsection{Counterfactual Reasoning Ability}
\begin{figure}[t!]
\centering
\includegraphics[width=0.99\textwidth]{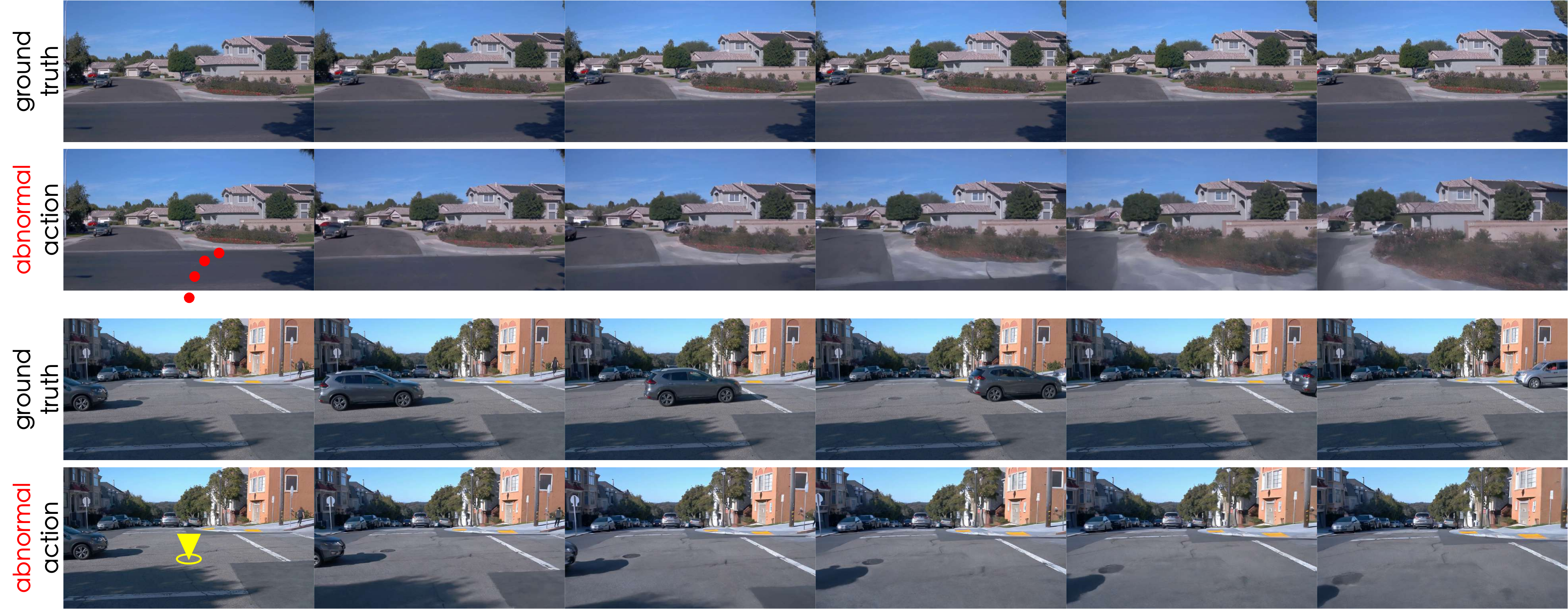}
\vspace{-2mm}
\caption{\textbf{Counterfactual reasoning ability.} By imposing actions that violate the traffic rules, we discover that \modelname can also predict the consequences of abnormal interventions. In the first example, the ego-vehicle passes over the road boundary and rushes into the bush following our instructions. In the second example, the passing car stops and waits to avoid a collision when we force the ego-vehicle to proceed at the crossroads. This showcases \modelname's potential for facilitating closed-loop simulation.}
\label{fig:counterfactual}
\vspace{-2mm}
\end{figure}

\vspace{-0.5mm}
Counterfactual reasoning ability is one of the emergent abilities of world models~\cite{gupta2024essential}. As shown in \cref{fig:counterfactual}, \modelname can effectively predict the counterfactual consequences caused by abnormal actions.

\subsection{Human Evaluation Cases}
\begin{figure}[t!]
\centering
\includegraphics[width=0.99\textwidth]{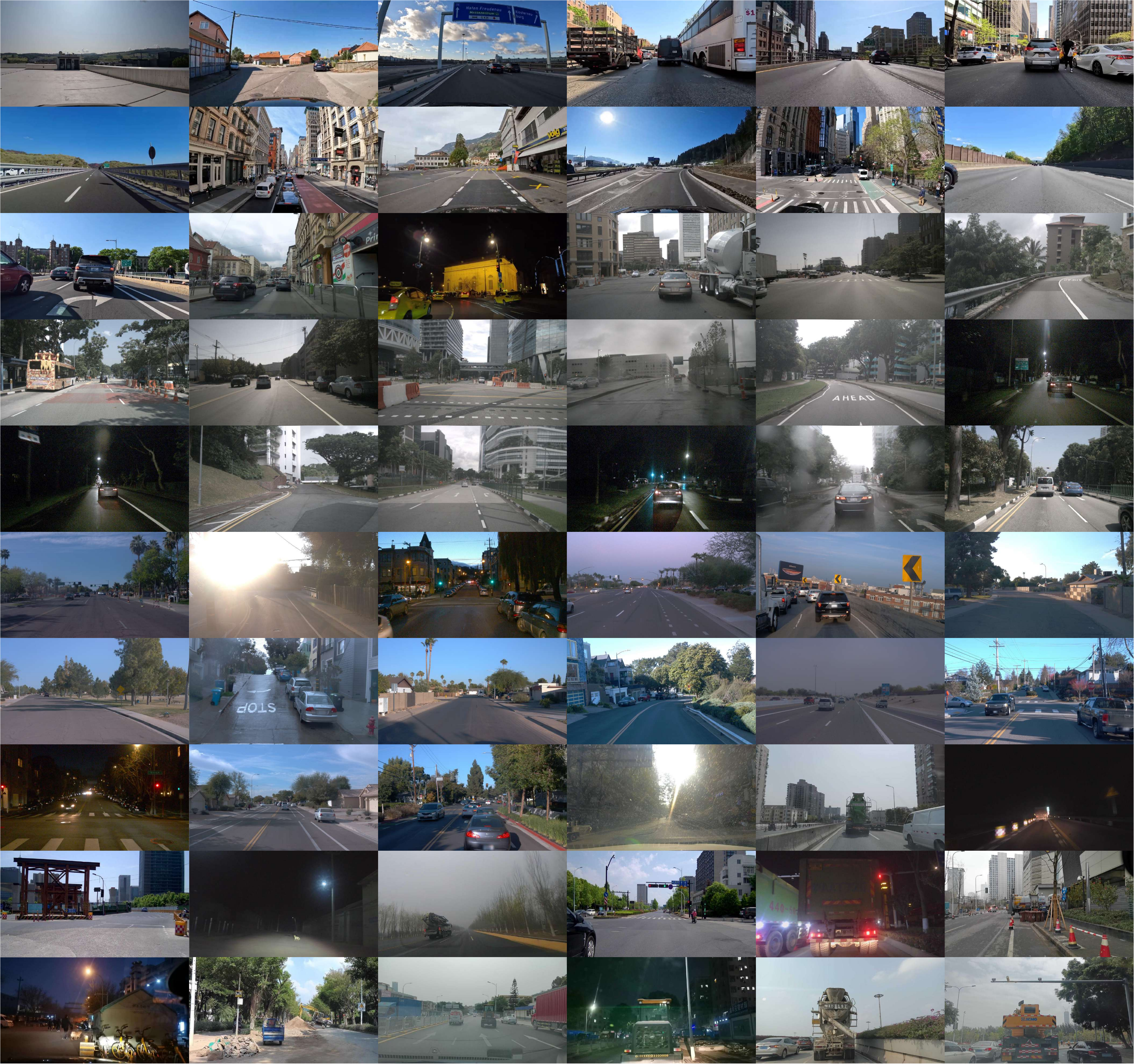}
\vspace{-2mm}
\caption{\textbf{Diverse scenes collected for human evaluation.} We carefully curate 60 scenes from OpenDV-YouTube-val, nuScenes, Waymo and CODA. The distinctive attributes of each dataset jointly represent the diversity of real-world environments, permitting a comprehensive human evaluation.}
\label{fig:collection}
\vspace{-2mm}
\end{figure}

\vspace{-0.5mm}
To demonstrate the diversity of the scenes selected for human evaluation (\cref{sec:experiments}), we show all cases gathered from OpenDV-YouTube-val~\cite{yang2024generalized}, nuScenes~\cite{caesar2020nuscenes}, Waymo~\cite{sun2020scalability}, and CODA~\cite{li2022coda} in \cref{fig:collection}.

\section{Licence of Assets}
\label{sec:license}
Our training and evaluation utilize the data from four publicly licensed datasets~\cite{caesar2020nuscenes,li2022coda,sun2020scalability,yang2024generalized}. Our implementation is based on the codebase of SVD~\cite{blattmann2023stable}, which uses the MIT license. The pretrained checkpoint of SVD is distributed under the stable video diffusion non-commercial community license.

\end{document}